\documentclass[sigconf]{acmart}
\AtBeginDocument{%
  \providecommand\BibTeX{{%
    \normalfont B\kern-0.5em{\scshape i\kern-0.25em b}\kern-0.8em\TeX}}}

\copyrightyear{2026}
\acmYear{2026}
\setcopyright{cc}
\setcctype{by}
\acmConference[SIGIR '26]{Proceedings of the 49th International ACM SIGIR Conference on Research and Development in Information Retrieval}{July 20--24, 2026}{Melbourne, VIC, Australia}
\acmBooktitle{Proceedings of the 49th International ACM SIGIR Conference on Research and Development in Information Retrieval (SIGIR '26), July 20--24, 2026, Melbourne, VIC, Australia}
\acmDOI{10.1145/3805712.3809655}
\acmISBN{979-8-4007-2599-9/2026/07}

\usepackage{amsfonts, amsmath, amsthm}
\usepackage[ruled, vlined, linesnumbered]{algorithm2e}
\usepackage{graphicx}
\usepackage{multirow}
\usepackage{nccmath}
\usepackage{subfig}
\usepackage{xcolor, xspace}
\usepackage{placeins}
\usepackage{pifont}
\usepackage{colortbl}
\usepackage{enumitem}
\usepackage{soul}
\usepackage{stfloats}  

\newtheorem{definition}{Definition}

\newcommand{\eg}{\textit{e.g.}\xspace}
\newcommand{\ie}{\textit{i.e.},\xspace}

\newcommand\figref[1]{Fig.~\ref{#1}}
\newcommand\algref[1]{Algorithm~\ref{#1}}
\newcommand\tabref[1]{Table.~\ref{#1}}
\newcommand\secref[1]{Sec.~\ref{#1}}
\newcommand\equref[1]{Eq.~(\ref{#1})}

\newcommand\appref[1]{Appendix~\ref{#1}}

\newcommand{\fakeparagraph}[1]{\noindent\textbf{#1.}}
\newcommand{\sysname}{FedMosaic\xspace}
\newcommand{\bluecomment}[1]{\textcolor{blue}{\texttt{// #1}}}
\newcommand{\graycomment}[1]{\textcolor{gray}{\texttt{// #1}}}

\ifodd 1

\else

\fi

\begin{document}

\title{\sysname: Federated Retrieval-Augmented Generation via Parametric Adapters}

\author{Zhilin Liang}
\affiliation{
  \department{SKLCCSE Lab}
  \institution{Beihang University}
  \city{Beijing}
  \country{China}
}
\email{zlliang@buaa.edu.cn}

\author{Yuxiang Wang}
\affiliation{
  \department{SKLCCSE Lab}
  \institution{Beihang University}
  \city{Beijing}
  \country{China}
}
\email{yuxiangwang@buaa.edu.cn}

\author{Zimu Zhou}
\affiliation{%
  \department{Department of Data Science} 
  \institution{City University of Hong Kong}
  \city{Hong Kong}
  \country{China}
}
\email{zimuzhou@cityu.edu.hk}

\author{Hainan Zhang}
\affiliation{
  \department{Beijing Advanced Innovation Center}
  \institution{Beihang University}
  \city{Beijing}
  \country{China}
}
\email{zhanghainan1990@163.com}

\author{Boyi Liu}
\affiliation{
  \department{SKLCCSE Lab}
  \institution{Beihang University}
  \city{Beijing}
  \country{China}
}
\email{boyliu@buaa.edu.cn}

\author{Yongxin Tong}
\authornote{Corresponding author.}
\affiliation{
  \department{SKLCCSE Lab}
  \institution{Beihang University}
  \city{Beijing}
  \country{China}
}
\email{yxtong@buaa.edu.cn}

\renewcommand{\shortauthors}{Zhilin Liang et al.}

\begin{abstract}
Retrieval-Augmented Generation (RAG) enhances Large Language Models (LLMs) by grounding generation in external knowledge to improve factuality and reduce hallucinations. 
Yet most deployments assume a centralized corpus, which is infeasible in privacy-aware domains where knowledge remains siloed. 
This motivates federated RAG (FedRAG), where a central LLM server collaborates with distributed silos without sharing raw documents.
In-context RAG violates this requirement by transmitting verbatim documents, whereas parametric RAG encodes documents into lightweight adapters that merge with a frozen LLM at inference, avoiding raw-text exchange. 
We adopt the parametric approach but face two unique challenges induced by FedRAG: high storage and communication from per-document adapters, and destructive aggregation caused by indiscriminately merging multiple adapters. 
We present \sysname, the first federated RAG framework built on parametric adapters.
\sysname clusters semantically related documents into multi-document adapters with document-specific masks to reduce overhead while preserving specificity, and performs selective adapter aggregation to combine only relevance-aligned, non-conflicting adapters.
Experiments show that \sysname achieves an average $10.9\%$ higher accuracy than 
state-of-the-art methods in four categories,
while lowering storage costs by $78.8\%$ to $86.3\%$ and communication costs by $91.4\%$, and never sharing raw documents\footnote{We have open-sourced the code at: https://github.com/lewellin727/FedMosaic}. 
\end{abstract}



\maketitle

\section{Introduction}
Retrieval-Augmented Generation (RAG) \cite{fan2024survey} has emerged a core technique for enhancing Large Language Models (LLMs) by grounding their responses in \textit{dynamic external knowledge} rather than static pre-training alone.
By retrieving relevant documents and conditioning generation on them, RAG improves factual accuracy and mitigates hallucination, driving its adoption in applications such as conversational search \cite{yao2023react}, enterprise knowledge assistants \cite{chen2025omniRAG}, and domain-specific Q\&A \cite{zhao2025medrag}. 
However, most RAG deployments assume access to a \textit{centralized} corpus (\eg Wikipedia, Common Crawl, or enterprise repositories). 
While feasible in open-domain settings, this assumption fails in vertical domains such as healthcare and finance, where critical knowledge remains locked in institutional silos due to privacy and compliance constraints.

\begin{figure}[t]
  \centering
    \includegraphics[width=0.47\textwidth]{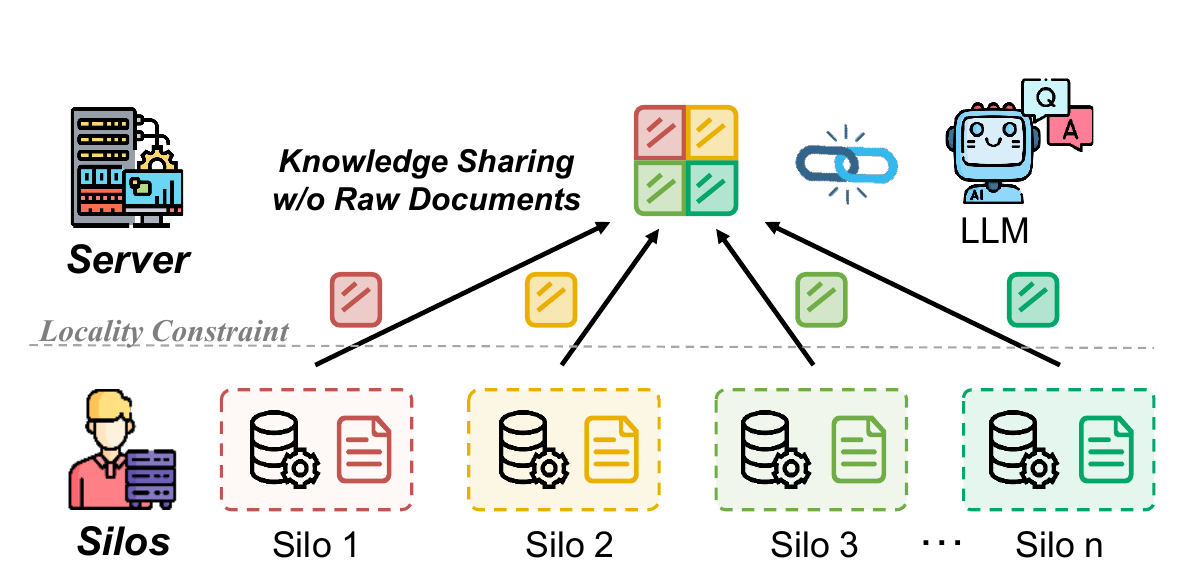}
    \caption{Federated RAG with locality constraint.}
  \label{fig:rag_vs_fedrag}
\end{figure}

This limitation motivates \textit{Federated RAG} (FedRAG), 
where a central LLM server collaborates with multiple distributed silos, each maintaining its own local corpus and serving as a knowledge provider without sharing its raw documents, \ie the \textit{locality constraint} (see \figref{fig:rag_vs_fedrag}).
In rare-disease diagnosis, for instance, a medical LLM could generate well-informed suggestions by leveraging clinical narratives (\eg pathology reports, triage notes, genomic records) from multiple hospitals.
No single hospital offers sufficient coverage for reliable decisions \cite{boycott2025rdi}, and regulations such as HIPAA \cite{US1996HIPAA} and GDPR \cite{EU2016GDPR} prohibit centralizing sensitive records \cite{shen2025rapid}. 
FedRAG allows the server to integrate knowledge across hospitals without exposing raw documents, supporting robust diagnosis while ensuring compliance.
More broadly, it extends RAG to distributed, privacy-conscious environments that increasingly characterize modern web-of-silos information systems.

Despite rapid progress in RAG, existing methods cannot be directly extended to the federated setting because they violate the locality constraint.
Conventional RAG depends on \textit{in-context} integration, where retrieved passages are inserted into the LLM’s prompt \cite{guu2020retrieval, lewis2020retrieval, dong2025understand, su2024dragin, yao2023react}. 
In a federated environment, this requires the server to fetch verbatim documents from silos, which inherently violates locality.
Other efforts pursue privacy-preserving prompt engineering by injecting noise into raw documents, but our empirical study shows that these approaches suffer severe accuracy degradation in FedRAG settings (see \secref{app:pppe}).
Noise injected independently by silos accumulates at the server, leading to performance even worse than single-silo retrieval.

Recent advances in parametric RAG \cite{su2025parametric} offer a promising alternative. 
Instead of raw text, documents are encoded into lightweight adapters (\eg LoRA) that can be merged with the frozen base LLM model at inference. 
It avoids raw-text exchange and naturally satisfies the locality constraint, making it an attractive candidate for FedRAG. 
In practice, each silo can encode its local documents into adapters and upload relevant ones to the server for adapter aggregation and response generation.
Yet directly extending parametric RAG to federated environments is far from trivial due to two challenges.
 \textit{(i) Storage and communication overhead}.
Training one adapter per document leads to an explosion in adapter count, imposing unsustainable storage demands at silos and heavy communication costs. 
A natural remedy is to encode \textit{multiple} documents per adapter, but naive grouping of heterogeneous documents causes \textit{intra-silo adapter interference}. 
\textit{(ii) Destructive adapter aggregation}:
As in federated learning, the standard adapter aggregation strategy is to average adapters across \textit{all} silos.
While this initially broadens knowledge coverage, \textit{indiscriminate} aggregation causes \textit{inter-silo adapter interference}.
Irrelevant adapters inject noise, and parameter averaging creates conflicts.

To address these challenges, we propose \sysname, an efficient and accurate federated RAG framework built upon parametric adapters. 
\sysname enforces the locality constraint while reducing overhead and mitigating interference through two innovations.
\textit{(i) Multi-Document Parametric Adapters}.
It clusters \textit{semantically related} documents into a single adapter while learning \textit{document-specific masks} to gate parameters. 
This preserves per-document specificity, prevents conflicting signals within an adapter, and substantially reduces storage and communication costs. 
\textit{(ii) Selective Adapter Aggregation}.
It ensures that only relevant and non-conflicting adapters are combined across silos. 
For each query, silos re-rank their local documents and share only relevance scores and masks.
The server then selects globally relevant adapters with minimal parameter conflict and aggregates them to generate responses.
Analogous to arranging tiles into a coherent mosaic, \sysname composes only the most relevant knowledge across silos, achieving locality-preserving, relevance-aware, and efficient federated RAG.

Our main contributions are summarized as follows.
\begin{itemize}
    \item 
    We introduce the first framework for federated RAG that enforces the locality constraint, enabling multi-silo retrieval without transmitting raw documents.
    \item
    We propose \sysname, a unified design that tackles the unique efficiency and accuracy challenges in federated parametric RAG. 
    It introduces multi-document parametric adapters with document-specific masks to reduce storage and communication overhead while preserving per-document specificity, and selective adapter aggregation to mitigate inter-silo conflicts by aggregating only relevant, non-overlapping adapters.
    \item 
    Extensive evaluations on four datasets show that \sysname achieves an average $10.9\%$ higher accuracy than state-of-the-art methods in four categories: local RAG \cite{wei2022chain, yao2023react, su2024dragin}, in-context FedRAG \cite{addison2024c, zhao2024frag, shojaee2025federated, guerraoui2025efficient}, federated fine-tuning \cite{zhang2024fedit, wang2024flora}, and parametric RAG \cite{su2025parametric}, while enforcing locality constraint and yielding a reduction of $78.8\%$to $86.3\%$ in storage and $91.4\%$ in communication cost.
\end{itemize}

\section{Preliminaries}
\label{sec:pre}

\subsection{Federated RAG}
\label{sec:pre:problem}
\fakeparagraph{Problem Setting}
Consider a federation with a central LLM server $\mathcal{G}$ and $M$ data silos $\{\mathcal{S}_1,\ldots,\mathcal{S}_M\}$ (\eg distinct organizations or geo-distributed data centers). 
Each silo $\mathcal{S}_m$ maintains a local knowledge corpus $\mathcal{D}_m = \{d^1_m, \ldots, d^{N_m}_{m}\}$, where each $d^i_{m}$ is a document. 
The silos act as distributed external knowledge providers, while the server is responsible for generating responses. 

Given a user query $q$, federated retrieval-augmented generation aims to produce an answer $a$ using knowledge across $\{\mathcal{D}_m\}_{m=1}^M$ under the \textit{locality constraint}: no plaintext document $d^i_m$ may leave its silo $\mathcal{S}_m$. 
The objective is to maximize answer \textit{accuracy} while keeping \textit{communication} overhead low.

\fakeparagraph{Limitations of Prior Arts}
Existing federated RAG frameworks \cite{wang2024feb4rag, shojaee2025federated, addison2024c,zhao2024frag, guerraoui2025efficient} rely on \textit{in-context} integration, where the server fetches \textit{verbatim} documents from silos and inserts them into the LLM context, which violates the \textit{locality constraint} by nature.

\subsection{Parametric Adapters}
\label{sec:pre:parametric}
Parametric RAG \cite{su2025parametric} offers a promising alternative by integrating external knowledge \textit{without} in-context documents.
It converts documents into \textit{parametric adapters} that directly augment the LLM. The pipeline consists of two steps. 
\begin{itemize}
    \item \textit{Document Augmentation.} 
    For each document $d^i$, an LLM generates $n$ rewrites and $m$ question-answer (QA) pairs, forming an augmented set
    \begin{equation}
        D^i = \{(d^{i, k}, q^{i, j}, a^{i, j}) \mid 1 \leq k \leq n,\, 1 \leq j \leq m\},
        \label{eq:augment}
    \end{equation}
    where $d^{i, k}$ is a rewrite and $(q^{i, j}, a^{i, j})$ is a QA pair.
    \item \textit{Parametric Encoding.}
    Using LoRA adapters \cite{hu2022lora}, each $D^i$ is encoded into a parametric form by minimizing the next-token loss:
    \begin{equation}
        \min_{\Delta\theta} \sum_{x \in D^i} \sum_{t=1}^T -\log P_{\theta + \Delta\theta}(x_t \mid x_{<t}),
        \label{eq:encode}
    \end{equation}
    where $\theta$ is the frozen LLM parameters, $\Delta\theta$ is the trainable LoRA parameters, $x_t$ is the $t$-th token in sequence $x$, and $T$ is the sequence length.
\end{itemize}
At inference, relevant adapters are composed with $\theta$ rather than sending raw text. 
This \textit{parametric} integration naturally enforces \textit{locality}, making it an attractive candidate for federated RAG. 
\algref{alg:fedprag} illustrates the straightforward extension of parametric RAG to federated scenarios.

\begin{algorithm}[t]
    \caption{Naive Federated Parametric RAG}
    \label{alg:fedprag}
    \KwIn{user query $q$, LLM server $\mathcal{G}$, data silos $\{\mathcal{S}_m\}_{m=1}^{M}$}
    \KwOut{answer $a$}

    \bluecomment{Offline Stage}
    
    \For{each silo $\mathcal{S}_m$}
    {
        Train an independent LoRA adapter for each document
    }

    \bluecomment{Online Stage}
    
    \For{each silo $\mathcal{S}_m$}
    {
        Retrieve top-$k$ documents based on $q$ and upload the corresponding averaged adapters to the server
    }
    Server aggregates multiple adapters by parameter summation and generates answer $a$
    
    \Return $a$
\end{algorithm}
\subsection{Challenges}
\label{sec:challenge}
\begin{figure}[t]
  \centering
  \subfloat[Grouped Documents]{
    \includegraphics[width=0.22\textwidth]{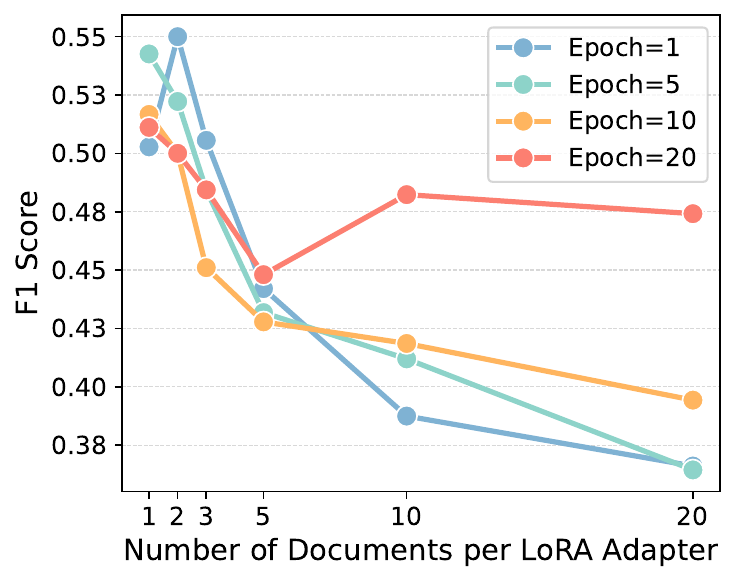}
    \label{fig:motivation-1}
  }
  \hfill
  \subfloat[Multiple LoRA Aggregation]{
    \includegraphics[width=0.22\textwidth]{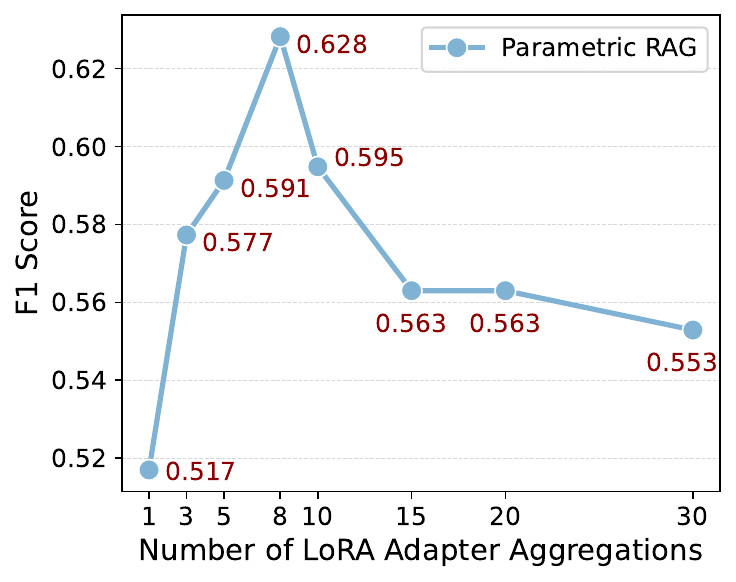}
    \label{fig:motivation-2}
  }
  \caption{Accuracy curves of parametric RAG for (a) grouped documents under a LoRA adapter across different training epochs, and (b) aggregation of multiple LoRA adapters.
  }
  \label{fig:motivation}
\end{figure}

Although parametric RAG is attractive, directly extending it to the federated setting introduces two challenges.
\begin{itemize}
    \item \textbf{Storage and Communication Overhead}.
    Training one adapter \textit{per document} causes an explosion in the number of adapters, leading to excessive storage requirements at silos and high silo-server communication costs.
    A natural remedy is to encode \textit{multiple documents} into a single adapter.
    Yet experiments show that naive grouping of heterogeneous documents severely degrades accuracy.
    As shown in \figref{fig:motivation-1}, when the number of documents per adapter increases from $1$ to $20$, the F1 score of parametric RAG at a single silo drops sharply\footnote{Results are measured on 2WikiMultihopQA Bridge dataset. We randomly group a fixed number of documents to each adapter at different epochs. At inference time, the response is generated using the adapter most relevant to the query.}. 
    The degradation persists across training epochs, indicating that heterogeneous documents within a shared adapter cause \textit{intra-silo adapter interference}.
    \item \textbf{Destructive Adapter Aggregation}.
    In federated RAG, a straightforward strategy is to aggregate adapters from \textit{all} silos by averaging their parameters. 
    While this can initially improve accuracy by incorporating \textit{diverse} knowledge, \textit{indiscriminate} aggregation ultimately harms accuracy. 
    As shown in \figref{fig:motivation-2}, the F1 score improves when a small number of relevant adapters are averaged, but declines as more are included\footnote{Results are measured on 2WikiMultihopQA Comparison dataset. Each document has an independent adapter trained for $3$ epochs. At inference time, we increase the number of averaged adapters for the same query from $1$ to $30$. }. 
    This degradation arises from \textit{inter-silo adapter interference}, caused by two factors: \textit{(i)} not all silo documents are equally relevant to a given query, so aggregating too many introduces noise, and \textit{(ii)} parameter averaging can create parameter conflicts among adapters \cite{yu2024dare, lu2024twin, huang2024emr, ortiz2023task}.  
    Together, these effects lead to destructive aggregation and reduced answer quality.
\end{itemize}

\fakeparagraph{Objectives}
These challenges motivate the design of a federated RAG framework that:  
\textit{(i)} preserves locality by exchanging only parametric adapters,  
\textit{(ii)} reduces storage and communication costs by supporting multi-document adapters without intra-silo interference, and  
\textit{(iii)} mitigates destructive aggregation through selective adapter combination that avoids inter-silo interference.  
These objectives underpin the design of \sysname.

\section{Method}
\label{sec:method}

\subsection{\sysname Overview}
\label{sec:method:overview}{We present \sysname, an efficient and accurate federated RAG framework built upon parametric adapters (see \figref{fig:overview}). 
\sysname addresses the challenges in \secref{sec:challenge} by \textit{(i)} encoding multiple semantically related documents into shared adapters, and \textit{(ii)} selectively aggregating relevant, non-conflicting adapter parameters across silos.  
Analogous to a mosaic that arranges relevant tiles into a coherent image, \sysname groups documents into parametric adapters and assembles only the most relevant ones across silos to support efficient and accurate federated RAG.
To our knowledge, \sysname is the first federated RAG scheme that ensures the \textit{locality constraint}. 

\begin{figure*}[t]
  \centering
    \includegraphics[width=0.96\textwidth]{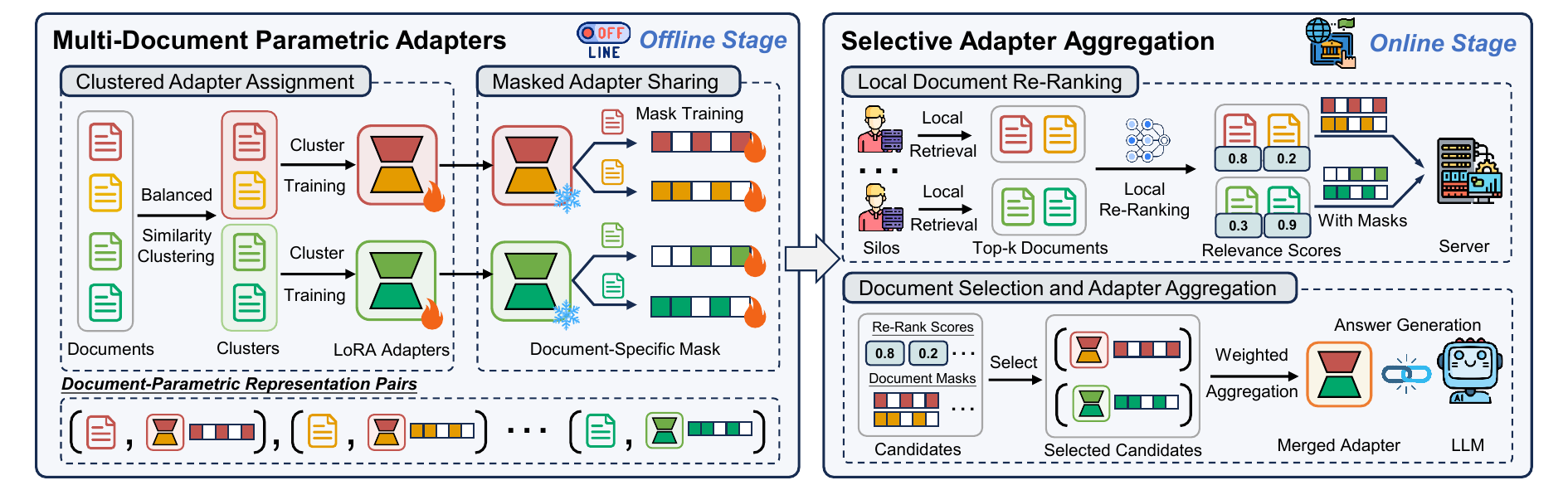}
    \caption{\sysname architecture and workflow.}
  \label{fig:overview}
\end{figure*}

\fakeparagraph{Architecture}
\sysname consists of two functional modules.
\begin{itemize}
    \item \textbf{Multi-Document Parametric Adapters} (\secref{sec:multi}).
    To reduce storage and communication overhead, \sysname shares one adapter across a cluster of semantically coherent documents. 
    This is feasible because a single document primarily activates a small subset of adapter parameters.
    Accordingly, \sysname \textit{(i)} groups semantically similar documents and trains one cluster-level adapter per cluster, and \textit{(ii)} learns document-specific masks that gate adapter parameters during adapter aggregation and answer generation. 
    This design preserves per-document specificity and mitigates intra-silo adapter interference during adapter sharing.
    \item \textbf{Selective Adapter Aggregation} (\secref{sec:selective}).
    To mitigate inter-silo adapter interference during adapter averaging, \sysname aggregates only adapters associated with the most relevant documents and least conflicting parameters.  
    This is enabled by the document-specific masks from the multi-document adapters. 
    For each query, silos re-rank their local documents and upload only the relevance scores and corresponding masks.    
    The server then selects the globally most relevant masks while minimizing parameter conflicts, measured by overlap among mask positions. 
    The corresponding adapters are requested, gated by the selected masks, and averaged using standard parameter aggregation.   
    This yields a relevance- and conflict-aware aggregation that preserves locality and improves answer quality.
\end{itemize}

\fakeparagraph{Workflow} 
We assume each silo has the same base LLM as the server and employs a homogeneous re-ranking model $\mathcal{M}_r$.  
These assumptions are standard in cross-silo federated learning, where participants typically have sufficient resources to fine-tune large models locally. 
Under this setting, \sysname operates in two stages.
\begin{itemize}
    \item \textbf{Offline Stage}.  
    Each silo clusters its local documents, trains a cluster-level adapter for each cluster, and learns a binary mask for every document.  
    The resulting adapters and masks are stored locally as the parametric representation of the silo’s corpus.
    \item \textbf{Online Stage}.  
    When a query arrives, the server broadcasts it to all silos.  
    Each silo performs local retrieval, applies the re-ranking model to compute relevance scores, and uploads the scores together with the masks of the retrieved documents.  
    The server then selects top-ranked documents, requests their adapters from silos, aggregates the masked adapters, and composes the result with the base LLM to generate the final answer.
\end{itemize}

\algref{alg:overall} illustrates the overall workflow.
In the \textit{offline} stage, each silo first performs balanced clustering on its local document corpus and trains an adapter for each cluster (line 4-5), then train a mask for each document (lines 7). 
This stage is executed only once.
In the \textit{online} stage, the server broadcasts the user query to all silos. 
Each silo retrieves top-$k$ documents and computes their relevance scores (lines 11-13). The server then selects documents based on the relevance scores and corresponding masks (lines 15-16), aggregates the associated LoRA adapters weighted by the scores, and generates the final answer $a$ (line 18-19).

{
\setlength{\textfloatsep}{0pt} 
\begin{algorithm}[t]
    \caption{\sysname}
    \label{alg:overall}
    \KwIn{user query $q$, LLM server $\mathcal{G}$, data silos $\{\mathcal{S}_m\}_{m=1}^{M}$}
    \KwOut{answer $a$}

    \bluecomment{Offline Stage}
    
    \For{each silo $\mathcal{S}_m$}
    {
        \graycomment{Clustered Adapter Assignment}
        
        Obtain clusters $\left\{ C_m^{1}, C_m^{2}, \dots, C_m^{t} \right\}$ via \equref{eq:clustering}
        
        Train adapters $\{(A_{m}^{i}, B_{m}^{i})\}_{i=1}^{t}$ per cluster via \equref{eq:encode}

        \graycomment{Masked Adapter Sharing}

        Train mask per document via \equref{eq:total_loss}
    }

    \bluecomment{Online Stage}

    \graycomment{Local Document Re-Ranking}
    
    \For{each silo $\mathcal{S}_m$}
    {
        Retrieve top-$k$ documents $\{d_{m}^{r_1}, \cdots, d_{m}^{r_k}\}$ based on $q$

        Compute relevance scores $\{s_{m}^{r_1}, \cdots, s_{m}^{r_k}\}$ via $\mathcal{M}_r$

        Upload the relevance scores and the corresponding document masks $\{(s_{m}^{r_i}, M_{m}^{r_i})\}_{i=1}^{k}$ to the server
    }

    \graycomment{Conflict-Aware Document Selection}
    
    Server selects $k^{'}$ documents via \equref{eq:selection} and notifies silos
    
    Silos upload corresponding LoRA adpters $\{(B_{i},A_{i})\}_{i=1}^{k^{'}}$

    \graycomment{Masked Adapter Aggregation}
    
    Server aggregates LoRA adapters $\Delta W_{\text{merge}}$ via \equref{eq:weighted}

    LLM generates answer $a \gets \mathcal{G}(q, \Delta W_{\text{merge}})$

    \Return $a$
\end{algorithm}
}

\subsection{Multi-Document Parametric Adapters}
\label{sec:multi}
To reduce silo-side storage and silo–server communication while avoiding \textit{intra-silo adapter interference}, \sysname replaces per-document adapters with \textit{multi-document} adapters. 
The design has two components: \textit{(i) clustered adapter assignment}, which groups semantically similar documents and trains one adapter per cluster (\secref{sec:clustering}); and \textit{(ii) masked adapter sharing}, which learns document-specific binary masks to activate distinct parameter subsets within a shared adapter (\secref{sec:mask}).
To our knowledge, this is the first \textit{multi-document} parametric RAG scheme that exploits the sparsity of LoRA adapters.

\subsubsection{Clustered Adapter Assignment}
\label{sec:clustering}
This module partitions a silo’s corpus into balanced clusters of semantically related documents and trains one adapter per cluster.

\fakeparagraph{Balanced Document Clustering}
Each document is embedded into a vector representation, and semantic similarity is measured in the embedding space using \eg Euclidean distance.  

To control \textit{intra-silo adapter interference}, we restrict the maximum number of documents assigned to each cluster.
This is because grouping semantically related documents improves adapter sharing, but overly large clusters reintroduce interference. 

In \sysname, we adopt the constrained $k$-means clustering algorithm \cite{bradley2000constrained} with an empirically chosen maximum cluster size ($5$ to $10$, see \secref{sec:overhead_reduction}).  
Consequently, each silo $\mathcal{S}_m$ obtains a balanced partition of its local corpus
\begin{equation}\label{eq:clustering}
C_{m} = \{ C_m^{1}, C_m^{2}, \dots, C_m^{t} \},
\end{equation}
where $t$ is the number of clusters, and $C_m^{i}$ represents the set of documents assigned to the $i$-th cluster.

\fakeparagraph{Cluster-Level Parametric Adapters}
After clustering, adapters are trained at the level of clusters rather than individual documents as in \secref{sec:pre:parametric}.
For each cluster $C_{m}^{i}$ in silo $\mathcal{S}_m$, we construct an augmented document set $D_{m}^{i}$ following \equref{eq:augment}.  
This set is encoded into a LoRA adapter $(A_{m}^{i}, B_{m}^{i})$ via \equref{eq:encode}, where $B_{m}^{i} \in \mathbb{R}^{d \times r}$ and $A_{m}^{i} \in \mathbb{R}^{r \times d}$ define the low-rank update $\Delta W = B_{m}^{i} A_{m}^{i}$, with $d$ representing the LLM layer size and $r \ll d$ the adapter rank. 
Thus, each silo $\mathcal{S}_m$ obtains $t$ cluster-level adapters $\{(A_{m}^{1}, B_{m}^{1}), \dots, (A_{m}^{t}, B_{m}^{t})\}$, reducing the number of adapters compared to per-document training while retaining semantic coherence within clusters.

\subsubsection{Masked Adapter Sharing}
\label{sec:mask}
This module aims to learn \textit{document-specific binary masks} that activate distinct subsets of a shared adapter, thereby reducing \textit{intra-silo adapter interference}.
The design is motivated by two observations: \textit{(i)} LoRA adapters contain redundancy \cite{xu2024winning}, and \textit{(ii)} fine-tuning typically affects only subsets of model parameters, which differ across tasks \cite{yu2024dare, du2025neural}.  
We hypothesize that document-specific knowledge can also be captured by distinct subsets of LoRA parameters. 
Accordingly, \sysname learns binary masks over a \textit{frozen} cluster-level adapter so that each document activates a tailored subspace.  
These masks also assist in selective adapter aggregation (\secref{sec:selective}).

\fakeparagraph{Adapter Mask Designation}
Let $(A_{m}^{i}, B_{m}^{i})$ be the LoRA adapter for cluster $C_{m}^{i}$, which encodes documents $\{d_{m}^{i_1},\ldots,d_{m}^{i_n}\}$. 
Applying a binary mask directly to the full update $\Delta W = B_{m}^{i}A_{m}^{i}\in\mathbb{R}^{d\times d}$ would require $O(d^2)$ mask parameters, which is both storage- and computation-intensive. 
Instead, we adopt a lightweight \textit{row-wise} masking scheme on $B_{m}^{i}\in\mathbb{R}^{d\times r}$, requiring only $O(d)$ parameters.  

For document $d_{m}^{i_j}$, we define a binary vector $M_{m}^{i_j}\in\{0,1\}^{d}$ and apply it row-wise to $B_{m}^{i}$:
\begin{equation}
\tilde{B}_{m}^{i_j} = M_{m}^{i_j} \circ B_{m}^{i},
\end{equation}
where $\circ$ denotes the Hadamard product with broadcasting across columns.
To stabilize the magnitude of masked updates, we introduce a rescale factor:
\begin{equation}
    \lambda_{m}^{i_j} = \tfrac{d}{\|M_{m}^{i_j}\|_1},
\end{equation}
and define the masked low-rank update as
\begin{equation}
    \Delta \tilde{W}_{m}^{i_j} = \lambda_{m}^{i_j} \cdot \tilde{B}_{m}^{i_j} A_{m}^{i}.
\end{equation}
This ensures stable masked updating while allowing each document to activate only a subset of adapter rows.

\fakeparagraph{Adapter Mask Training}
During training, the adapter parameters $(A_{m}^{i},B_{m}^{i})$ remain frozen and only the masks are optimized.  
This preserves shared cluster-level knowledge while enabling each document to learn a sparse, specialized activation pattern. 

To enable gradient-based optimization, the binary mask is relaxed via a sigmoid parameterization $\sigma$.  
Let $\hat{M}_{m}^{i_j}\in\mathbb{R}^{d}$ be trainable logits.
During training we use $\sigma(\alpha\hat{M}_{m}^{i_j})$ with sharpening factor $\alpha>0$, and impose $\ell_1$ sparsity,
The total loss function is defined as
\begin{equation}
\label{eq:total_loss}
\min_{\hat M_{m}^{i_j}} \; 
\mathcal{L}(\hat M_{m}^{i_j}; \boldsymbol\theta_F) =
\mathcal{L}_{\mathrm{next}} (\hat M_{m}^{i_j}; \boldsymbol\theta_F)
+ 
\lambda_{\ell_1} \cdot \left\| \sigma\left(\alpha \cdot \hat M_{m}^{i_j}\right) \right\|_1
\;,
\end{equation}
where $\boldsymbol\theta_F$ is the frozen base and adapter parameters, $\mathcal{L}_{\mathrm{next}}$ is the next-token loss, and $\lambda_{\ell_1}$ balances task accuracy and mask sparsity.  

After training, the masks are binarized via thresholding:
\begin{equation}
{M}_{m}^{i_j} = \mathbb{I}\left[\hat M_{m}^{i_j} > 0\right],
\end{equation}
where $\mathbb{I}[\cdot]$ is the indicator function. 

This procedure is applied independently for each document in the cluster, producing a set of sparse, document-specific masks $\left\{M_{m}^{i_j}\right\}_{j=1}^{n}$ that \textit{(i)} mitigate intra-adapter interference by isolating document-specific subspaces, and \textit{(ii)} provide conflict indicators for selective aggregation across silos.

To further reduce the storage overhead introduced by the binary mask parameters, we adopt a bit-packing scheme that encodes each $d$-dimensional binary mask vector $M_{m}^{i_j} \in \{0,1\}^{d}$ into $\lceil d/8 \rceil$ bytes by packing 8 bits per byte.
\begin{equation}
\mathbf{b}_i = \sum_{j=1}^{8} m_{8(i-1)+j} \cdot 2^{j-1}, 
\quad \text{for } i = 1, \ldots, \left\lceil \frac{d}{8} \right\rceil, 
\end{equation}
where $\mathbf{b}_i$ denotes the $i$-th packed byte and $m_{8(i-1)+j}$ denotes the $8(i-1)+j$-th bit in $M_{m}^{i_j}$.
This reduces storage by 8$\times$ over naive byte-aligned \texttt{uint8} representations and supports encoding and decoding in $O(d)$ time via simple linear scans of the mask entries.

\subsection{Selective Adapter Aggregation}
\label{sec:selective}
As discussed in \secref{sec:challenge}, naive aggregation of adapters from all silos introduces \textit{inter-silo adapter interference}, since irrelevant documents introduce noise and overlapping parameters cause conflicts. 
To overcome this, \sysname introduces \textit{selective adapter aggregation}, which aggregates only query-relevant documents while suppressing parameter conflicts.  
The process consists of three steps: 
\textit{(i) local document re-ranking}, each silo re-ranks its local documents for the query and uploads only their relevance scores and masks (\secref{sec:rerank}); 
\textit{(ii) conflict-aware document selection}, the server selects a globally relevant subset under a conflict-aware criterion (\secref{sec:selection}); and 
\textit{(iii) masked adapter aggregation}, the selected adapters are aggregated in a mask-gated, relevance-weighted manner (\secref{sec:aggregation}). 
These designs address the unique challenges on accuracy when adapting parametric RAG to federated scenarios.



\subsubsection{Local Document Re-Ranking} 
\label{sec:rerank}
Upon receiving a query, the server broadcasts it to all silos. 
Each silo $\mathcal{S}_m$ retrieves $k$ candidate documents $\{d_{m}^{r_1},\dots,d_{m}^{r_k}\}$ and applies a re-ranking model $\mathcal{M}_r$, which is the same across silos, to assign normalized relevance scores $\{s_{m}^{r_1},\dots,s_{m}^{r_k}\}$.
For each candidate, the silo collects its associated mask $M_{m}^{r_i}$ over the cluster-level adapter and uploads only the pairs $\{(s_{m}^{r_i},M_{m}^{r_i})\}_{i=1}^{k}$, while keeping the raw documents and LoRA adapters local.


\subsubsection{Conflict-Aware Document Selection}
\label{sec:selection}
The server collects the relevance scores and masks of all $kM$ candidates and selects a subset $S$ of $k'$ documents that maximize relevance while minimizing parameter conflicts among their masks, given that relevance-based re-ranking alone is insufficient for RAG \cite{lee2025shifting}.
We measure conflicts between two masks $M_i$ and $M_j$ as their overlap:
\begin{equation}
    \text{overlap}(M_i,M_j) = \frac{\langle M_i,M_j \rangle}{d},
\end{equation}
where $d$ is the mask dimension.

The selection objective is formulated as
\begin{equation}\label{eq:selection}
\max_{S \subseteq [n],\, |S| = k'} \left( \sum_{i \in S} \frac{s_i}{k^{'}} - 2 \lambda_\text{ol} \cdot \frac{\sum_{\substack{i, j \in S, i < j}} \text{overlap}(M_i, M_j)}{k^{'} \cdot (k^{'}-1)} \right),
\end{equation}
where $[n]$ is the index set of all candidate documents, $s_i$ is the relevance score of document $i$, $M_i$ is its mask, and $\lambda_\text{ol} > 0$ balances relevance against conflicts. We prove the above selection problem is NP-hard in \appref{sec:np_hard}.

We solve \equref{eq:selection} via a greedy algorithm.
At each iteration, we select the candidate with the highest marginal gain
\begin{equation}
    \Delta_{i} = s_i - \lambda_\text{ol} \cdot \frac{1}{|S|} \sum_{j \in S} \text{overlap}(M_i, M_j),
\end{equation}
until $k'$ documents are chosen.
To avoid noisy selections, a score threshold $\tau$ is applied.
Only candidates with $s_i > \tau$ are considered, and the procedure terminates early if none remain.

\subsubsection{Masked Adapter Aggregation}
\label{sec:aggregation}
After selection, the server requests only the necessary information from silos. 
For each chosen document $i \in S$, this includes its mask $M_i$ and the associated cluster adapter $(B_{i},A_{i})$. 
Let $w_i = s_i / \sum_{j \in S} s_j$ be normalized relevance weights. 
The aggregated LoRA adapter is then computed as
\begin{equation}
\Delta W_{\text{merge}} = \sum_{i\in S} w_i \,(M_i \circ B_{i}) A_{i},
\label{eq:weighted}
\end{equation}
which is combined with the base LLM $\mathcal{G}$ to generate the final answer. 
This aggregation strategy ensures query relevance, suppresses parameter conflicts, and maintains locality of raw documents.

\section{Experiments}
\label{sec:exp}
\begin{table*}[!tbp]
\caption{Overall accuracy measured by F1 score for \sysname and four types of baseline methods.}
\label{tab:acc}
\begin{tabular}{@{}llcccccccc@{}}
\toprule
\multicolumn{1}{l}{\multirow{2}{*}{Type}} &
  \multicolumn{1}{l}{\multirow{2}{*}{Method}} &
  \multicolumn{2}{c}{HotpotQA} &
  \multicolumn{4}{c}{2WikiMultihopQA} &
  PopQA &
  CWQ \\ \cmidrule(l){3-10} 
\multicolumn{1}{c}{} &
  \multicolumn{1}{c}{} &
  Bridge &
  \multicolumn{1}{c}{Compare} &
  Bridge &
  Compare &
  Inf. &
  \multicolumn{1}{c}{Compose} &
  \multicolumn{1}{c}{Total} &
  Total \\ \midrule
\multicolumn{1}{l}{\multirow{4}{*}{Local}} &
  \multicolumn{1}{l}{Standard RAG} &
  0.1474 &
  0.3602 &
  0.2838 &
  0.2837 &
  0.1549 &
  0.0748 &
  0.1366 &
  0.2804 \\
\multicolumn{1}{l}{} &
  \multicolumn{1}{l}{CoTRAG \cite{wei2022chain}} &
0.0808	&
0.2632	&
0.3821	&
0.3444	&
0.1380	&
0.0377	&
0.0444	&
0.2378  \\
\multicolumn{1}{l}{} &
  \multicolumn{1}{l}{React \cite{yao2023react}} &
\underline{0.1731}	&
0.3103	&
0.3108	&
0.2608	&
0.1528	&
0.0590	&
0.1135	&
0.2443  \\

\multicolumn{1}{l}{} &
  \multicolumn{1}{l}{Dargin \cite{su2024dragin}} &
    0.1257	&
    0.3540	&
    0.3271	&
    0.3786	&
    0.1473	&
    0.0766	&
    0.0895	&
    0.3662 \\ \midrule
    
\multicolumn{1}{l}{\multirow{3}{*}{FedRAG}} &
  \multicolumn{1}{l}{FRAG \cite{addison2024c, zhao2024frag}} &
0.1317	&
0.3390	&
0.3357	&
0.2964	&
0.1527	&
0.0265	&
0.1207	&
0.1946  \\

\multicolumn{1}{l}{} &
  \multicolumn{1}{l}{MKPQA \cite{shojaee2025federated}} &
0.1562	&
0.2757	&
0.2831	&
0.2679	&
0.2112	&
0.0510	&
0.1329	&
0.2008  \\
\multicolumn{1}{l}{} &
  \multicolumn{1}{l}{RAGRoute \cite{guerraoui2025efficient}} &
0.1137	&
0.2825	&
0.2239	&
0.2472	&
0.1351	&
0.0293	&
0.0380	&
0.1767  \\ \midrule

\multicolumn{1}{l}{\multirow{2}{*}{FedFT}} &
  \multicolumn{1}{l}{FedIT \cite{zhang2024fedit}} &
  0.1107 &
  0.4188 &
  \underline{0.4250} &
  0.4377 &
  0.1996 &
  0.0692 &
  \underline{0.2152} &
  \underline{0.3687} \\
\multicolumn{1}{l}{} &
  \multicolumn{1}{l}{FLoRA \cite{wang2024flora}} &
  0.1030 &
  \underline{0.4259} &
  0.3625 &
  0.4038 &
  0.1864 &
  \underline{0.0770} &
  0.1917 &
  0.3497 \\ \midrule
\multicolumn{1}{l}{Param.} &
  \multicolumn{1}{l}{PRAG \cite{su2025parametric}} &
  0.0983 &
  0.4161 &
  0.3875 &
  \underline{0.4805} &
  \underline{0.2138} &
  0.0462 &
  0.0759 &
  0.2810 \\ \midrule
\multicolumn{1}{l}{Ours} &
  \multicolumn{1}{l}{\sysname} &
  \textbf{0.1980} &
  \textbf{0.4547} &
  \textbf{0.4453} &
  \textbf{0.5275} &
  \textbf{0.2474} &
  \textbf{0.0940} &
  \textbf{0.2368} &
  \textbf{0.3841} \\ \bottomrule
\end{tabular}
\end{table*}
\subsection{Experimental Setup}
\label{sec:exp:setup}

\fakeparagraph{Baselines} 
We compare \sysname with four categories of methods: 
local RAG (Standard RAG, CoTRAG \cite{wei2022chain}, ReAct \cite{yao2023react}, Dargin \cite{su2024dragin}), 
in-context FedRAG (FRAG \cite{addison2024c, zhao2024frag}, MKPQA \cite{shojaee2025federated}, RAGRoute \cite{guerraoui2025efficient}), 
federated fine-tuning (FedIT \cite{zhang2024fedit}, FLora \cite{wang2024flora}), and 
parameterized RAG (PRAG \cite{su2025parametric}). 
All baselines are adapted to the federated setting for fair comparison.
Methods based on privacy-preserving prompt engineering are not included as baselines due to their severe accuracy degradation (see \secref{app:pppe}).

\fakeparagraph{Datasets and Models} 
We experiment on HotpotQA (HQA) \cite{yang2018hotpotqav}, 2WikiMultihopQA (2WQA) \cite{ho2020wikiqa}, PopQA (PQA) \cite{mallen2023popqa}, and ComplexWebQuestions (CWQ) \cite{talmor2018complexqa} with different question types (\eg Bridge). 
By default, each dataset is subsampled to 300 Q\&A instances and a document corpus is constructed following \cite{su2025parametric}. This corpus is then partitioned into silos based on underlying topics using a Dirichlet-based allocation strategy with $\alpha=0.1$.
We also conduct privacy evaluations on Enron Emails and WikiText datasets following \cite{zeng2024good, zeng2024mitigating}.
LLaMA3.2-1B-Instruct and LLaMA3-8B-Instruct are used as backbone LLMs. 

\fakeparagraph{Metrics}
We report four metrics:  
(1) \textit{Accuracy}: correctness of server LLM responses to user queries measured by F1 score;  
(2) \textit{Privacy}: average success protection rate against targeted attack \cite{zeng2024good} and prefix attack \cite{carlini2022quantifying} at server-side inference;  
(3) \textit{Communication Efficiency}: average parameters transmitted per query from silo to server;  
(4) \textit{Storage Overhead}: extra silo-side storage for parametric adapters.


\subsection{Main Results}
\label{sec:main_result}
\tabref{tab:acc} reports the overall F1 score of \sysname and baselines on four datasets with the LLaMA3.2-1b-instruct backbone. 
Local RAG methods show unstable performance, since each silo can only rely on its own documents without cross-silo knowledge. 
In-context FedRAG baselines attempt to integrate cross-silo information but violate the locality constraint and often introduce noisy documents, limiting their gains. 
FedIT and FLoRA achieve relatively high scores on some subsets (\eg $0.4250$ on 2WQA Bridge) and exhibit more stable performance overall, but require $10\times$ higher training budgets.
Naive federated parametric RAG performs well on certain subsets (\eg $0.4805$ on 2WQA Compare), but suffers severe drops on others (\eg $0.2810$ on CWQ), consistent with the challenges in \secref{sec:challenge}.

In contrast, \sysname consistently achieves state-of-the-art performance across all datasets, demonstrating its effectiveness in mitigating both intra- and inter-silo adapter interference. 
Specifically, compared to the strongest baseline, \sysname improves the average F1 scores on HQA, 2WQA, and PQA by $10.57\%$, $13.08\%$, and $10.03\%$, respectively.
These results highlight the robustness of \sysname and its ability to leverage distributed knowledge while preserving locality constraints.

\begin{table}[t]
\caption{Privacy evaluation over target and prefix attack.}
\label{tab:private}
\begin{tabular}{@{}llcccc@{}}
\toprule
Attack &
  Method &
  \begin{tabular}[c]{@{}c@{}}Target\\ Prompts\end{tabular} &
  \begin{tabular}[c]{@{}c@{}}Target\\ Info.\end{tabular} &
  \begin{tabular}[c]{@{}c@{}}Repeat\\ Prompts\end{tabular} &
  \begin{tabular}[c]{@{}c@{}}Repeat\\ Length\end{tabular} \\ \midrule
\multirow{2}{*}{Target} & IC-FedRAG    & 138.93     & 102.20 & 131.33 & 21.34  \\
                        & \sysname  & \textbf{0}      & \textbf{0}      & \textbf{0}      & \textbf{0}      \\ \midrule
\multirow{2}{*}{Prefix} & IC-FedRAG    & 52,86  & 46.66  & 55.93  & 23.58  \\
                        & \sysname  & \textbf{44.00}  & \textbf{29.53}  & \textbf{22.86}  & \textbf{22.22}  \\ \bottomrule
\end{tabular}
\end{table}
\begin{table}[t]
\caption{LLaMA3-8B compared with competitive baselines.}
\label{tab:acc_8b}
\begin{tabular}{@{}llcccc@{}}
\toprule
\multicolumn{1}{l}{\multirow{2}{*}{Type}} & \multicolumn{1}{l}{\multirow{2}{*}{Method}} & \multicolumn{4}{c}{2WikiMultiHopQA} \\ \cmidrule(l){3-6} 
\multicolumn{1}{c}{}       & \multicolumn{1}{c}{}           & Bridge             & Compare             & Inf.               & Compose         \\ \midrule
\multicolumn{1}{l}{Local}  & \multicolumn{1}{l}{ReAct}     & 0.3432             & 0.4022              & 0.2093             & 0.0746    \\
\multicolumn{1}{l}{FedRAG} & \multicolumn{1}{l}{MKPQA}     & 0.2238             & 0.4011              & \underline{0.2797} & 0.0932  \\
\multicolumn{1}{l}{FedFT}  & \multicolumn{1}{l}{FedIT}     & \underline{0.4995} & \underline{0.6105}  & 0.2677             & \underline{0.1311}  \\
\multicolumn{1}{l}{FedFT}  & \multicolumn{1}{l}{FLoRA}     & 0.4701             & 0.5714              & 0.2355             & 0.1138  \\
\multicolumn{1}{l}{Param.} & \multicolumn{1}{l}{PRAG}      & 0.1812             & 0.2527              & 0.0829             & 0.0271  \\ \midrule
\multicolumn{1}{l}{Ours}   & \multicolumn{1}{l}{\sysname}  & \textbf{0.5207}    & \textbf{0.6237}     & \textbf{0.2859}    & \textbf{0.1584}  \\ \bottomrule
\end{tabular}
\end{table}

\subsection{Micro-Benchmarks}
\subsubsection{Privacy Evaluation} 
The experiment justifies the necessity of the locality constraint for FedRAG, and further quantifies privacy benefits of \sysname under the locality constraint versus in-context FedRAG.
We consider two data extraction attacks: \textit{targeted} and \textit{prefix} \cite{zeng2024good, carlini2022quantifying}. 
We construct $300$ privacy-sensitive samples following \cite{zeng2024mitigating}, and then apply retrieval-data attacks to in-context FedRAG and training-data attacks to \sysname. 
The in-context FedRAG (\ie IC-FedRAG) adopts a minimal FRAG \cite{addison2024c, zhao2024frag} design, capturing key characteristics of MKPQA \cite{shojaee2025federated} and RAGRoute \cite{guerraoui2025efficient}.

\tabref{tab:private} reports the attack success rates (lower is better) of \sysname and in-context FedRAG, showing that \sysname provides stronger resistance to data extraction attacks. 
Under targeted attacks, \sysname is almost immune ($0\%$ across all metrics) since the absence of explicit context makes it difficult to trigger adapter-encoded knowledge, whereas in-context FedRAG prompts are more easily repeated by the LLM. 
Under prefix attacks, \sysname also outperforms in-context FedRAG (\eg a $59\%$ reduction in repeated prompts and a $36.7\%$ reduction in target information), benefiting from training on rewritten data (see \equref{eq:augment}), where sensitive content can be filtered using simple prompts (\eg Ensure the revision has no emails). 
Furthermore, successful attacks require knowledge of the actual training data, which further increases attack difficulty and demonstrates the higher degree of privacy achieved by \sysname.

\subsubsection{Performance with Larger Backbones}
To evaluate the scalability of \sysname, we extend the backbone to LLaMA3-8B-Instruct. For a comprehensive comparison, we select the most competitive baselines reported in \secref{sec:main_result} and conduct experiments on the 2WQA dataset, as it contains the most diverse set of Q\&A types.

\tabref{tab:acc_8b} reports the F1 scores of \sysname and the competitive baselines. 
We observe that PRAG exhibits notable performance degradation due to accumulated noise and parameter conflicts. 
In contrast, \sysname consistently maintains state-of-the-art performance, demonstrating strong scalability to larger model sizes. 
For example, on the Bridge and Compose datasets, \sysname outperforms the strongest baselines by $4.2\%$ and $20.8\%$, respectively.

\begin{figure}[t]
  \centering
  \subfloat[Per-Document Storage]{
    \includegraphics[width=0.225\textwidth]{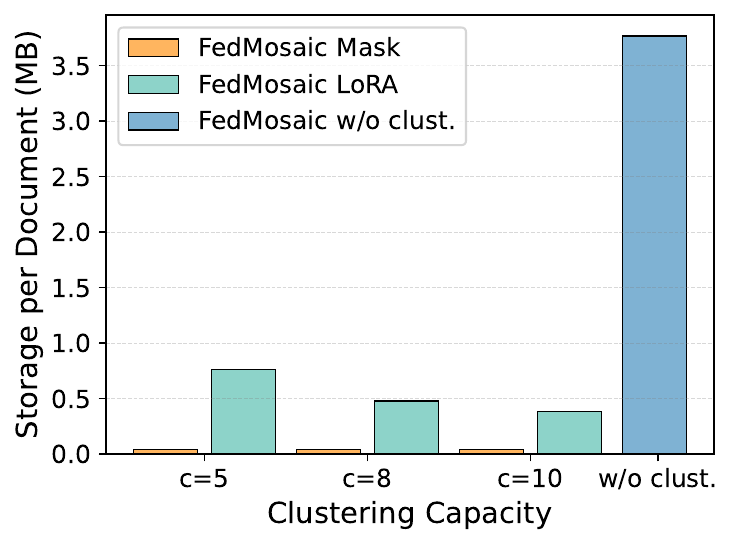}
    \label{fig:storage}
  }
  \hfill
  \subfloat[Per-Query Communication]{
    \includegraphics[width=0.225\textwidth]{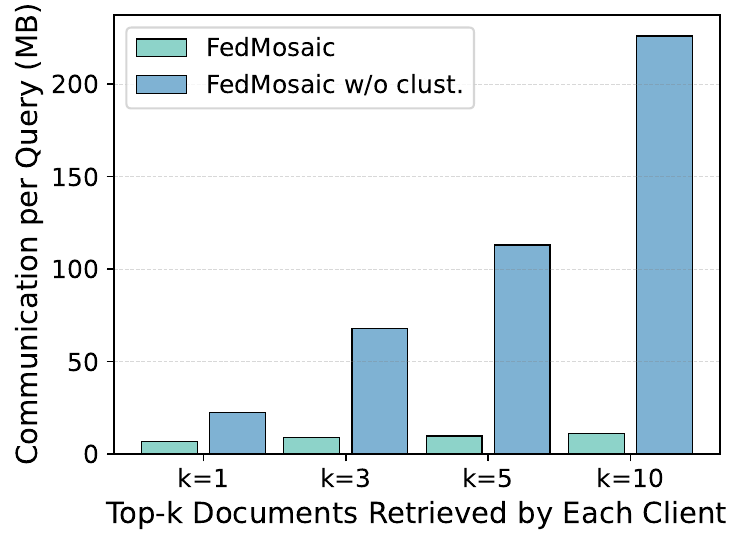}
    \label{fig:communication}
  }
  \vspace{-0.5em}
  \caption{Storage and communication overhead.}
  \vspace{-1em}
  \label{fig:storage+comm}
\end{figure}
\begin{table*}[t]
\caption{Performance of privacy-preserving prompt engineering methods.}
\label{tab:pppe}
\begin{tabular}{@{}lcccccccc@{}}
\toprule
\multirow{2}{*}{Method} & \multicolumn{2}{c}{HotpotQA} & \multicolumn{4}{c}{2WikiMultihopQA} & PopQA & CWQ \\ \cmidrule(l){2-9} 
             & Bridge & Compare & Bridge & Compare & Inf. & Compose & Total  & Total  \\ \midrule
DP\_Prompt   & $0.0228_{\downarrow 88.5\%}$ & $0.2179_{\downarrow 52.1\%}$ & $0.0785_{\downarrow 82.4\%}$ & $0.1305_{\downarrow 75.3\%}$ & $0.0385_{\downarrow 84.4\%}$ & $0.0114_{\downarrow 87.9\%}$ & $0.0060_{\downarrow 97.5\%}$ & $0.0778_{\downarrow 79.7\%}$ \\
Sage (Attr)  & $0.0791_{\downarrow 60.1\%}$ & $0.3286_{\downarrow 27.7\%}$ & $0.1808_{\downarrow 59.4\%}$ & $0.2218_{\downarrow 58.0\%}$ & $0.1318_{\downarrow 46.7\%}$ & $0.0222_{\downarrow 76.4\%}$ & $0.0676_{\downarrow 71.5\%}$ & $0.1217_{\downarrow 68.3\%}$ \\
Sage (Agent) & $0.0923_{\downarrow 53.4\%}$ & $0.3314_{\downarrow 27.1\%}$ & $0.2898_{\downarrow 34.9\%}$ & $0.2645_{\downarrow 49.9\%}$ & $0.1343_{\downarrow 45.7\%}$ & $0.0231_{\downarrow 75.4\%}$ & $0.0560_{\downarrow 76.4\%}$ & $0.1485_{\downarrow 61.3\%}$ \\
AUGPE        & $0.0079_{\downarrow 96.0\%}$ & $0.0514_{\downarrow 88.7\%}$ & $0.0153_{\downarrow 96.6\%}$ & $0.0283_{\downarrow 94.6\%}$ & $0.0225_{\downarrow 90.9\%}$ & $0.0032_{\downarrow 96.6\%}$ & $0.0004_{\downarrow 99.8\%}$ & $0.0247_{\downarrow 93.6\%}$ \\ \bottomrule
\end{tabular}
\end{table*}
\subsubsection{Compared to Privacy-Preserving Prompt Engineering Methods}
\label{app:pppe}
Privacy-preserving prompt engineering seeks to protect sensitive information by introducing noise through anonymization or differential privacy. Representative approaches include differential privacy-based methods (DP-Prompt \cite{utpala2023dpprompt}, AUGPR \cite{xie2024differentially}) and anonymization-based methods (Sage \cite{zeng2024mitigating}).

\tabref{tab:pppe} reports the F1 scores of these methods, together with their performance drops relative to \sysname. 
All methods show severe accuracy degradation, especially differential privacy approaches (\eg AUGPR), which perturb the global distribution by introducing noise that obscures document-specific information, and, even when applied document-wise (\eg DP-Prompt), can damage fine-grained knowledge crucial for accurate RAG.
Similarly, anonymization-based methods such as Sage also lose substantial information, with an average $53\%$ degradation, indicating that key evidence for Q\&A is often compromised during anonymizatio. 
Overall, the poor utility of these approaches in RAG tasks makes them unsuitable as primary baselines in our evaluation. 



\begin{table*}[t]
\caption{Accuracy improvement and robustness of selective aggregation under different selection $k$.}
\label{tab:agg}
\renewcommand{\arraystretch}{0.9} 
\setlength{\tabcolsep}{5pt}       
\begin{tabular}{@{}llccccccc@{}}
\toprule
Dataset & Type          & w/o Sel. & $k$=1  & $k$=3  & $k$=5  & $k$=10 & $k$=15 & $k$=30 \\ \midrule
\multirow{4}{*}{2WQA}   & Bridge   & $0.4224$   & $0.4552_{\uparrow 7.77\%}$      & $0.4447_{\uparrow 5.28\%}$ & $0.4492_{\uparrow 6.34\%}$ & $0.4525_{\uparrow 7.13\%}$ & $0.4525_{\uparrow 7.13\%}$ &                                  $0.4525_{\uparrow 7.13\%}$ \\
                        & Compare  & $0.5001$   & $0.4940_{\downarrow 1.22\%}$        & $0.5108_{\uparrow 2.14\%}$ & $0.5142_{\uparrow 2.82\%}$ & $0.5142_{\uparrow 2.82\%}$ & $0.5142_{\uparrow 2.82\%}$ & $0.5142_{\uparrow 2.82\%}$ \\
                        & Inf.     & $0.2137$   & $0.2316_{\uparrow 8.38\%}$        & $0.2581_{\uparrow 20.7\%}$ & $0.2455_{\uparrow 14.8\%}$ & $0.2445_{\uparrow 14.4\%}$ & $0.2422_{\uparrow 13.3\%}$ & $0.2422_{\uparrow 13.3\%}$ \\
                        & Compose  & $0.0862$   & $0.0750_{\downarrow 12.9\%}$        & $0.0869_{\uparrow 0.81\%}$ & $0.0953_{\uparrow 10.5\%}$ & $0.0953_{\uparrow 10.5\%}$ & $0.0953_{\uparrow 10.5\%}$ & $0.0953_{\uparrow 10.5\%}$ \\ \bottomrule
\end{tabular}
\end{table*}

\subsection{Ablation Studies}
We conduct a thorough ablation study to isolate and verify the contribution of each component in \sysname.
Specifically, we analyze \textit{multi-document parametric adapters} in terms of overhead reduction (\secref{sec:overhead_reduction}) and mask effectiveness (\secref{sec:contributions_of_mask}), while \textit{selective adapter aggregation} is evaluated for the impact of the selection $k$ (\secref{sec:effectiveness_of_aggregation}) and the retrieval $k$ (\secref{sec:retrieval_k}) on accuracy.

\subsubsection{Overhead Reduction by Clustering}
\label{sec:overhead_reduction}

\figref{fig:storage+comm} compares the storage and communication overhead of \sysname with the w/o clustering variant.
We choose $C \in \{5, 8, 10\}$ and $k \in \{1, 3, 5, 10\}$ for silo retrieval in our experiments.
\figref{fig:storage} shows that adapter size inversely proportional decreases with clustering (\eg $12.67\%$ of no-clustering at $c=8$), while per-document masks stay constant at $\sim1.03\%$.
The total storage is reduced to $11.23\%$ of the w/o clustering variant at $c=10$.
\figref{fig:communication} shows that per-query communication increases linearly in the w/o clustering variant, whereas clustering keeps it stable; at $k=10$, the cost is reduced to $4.86\%$ of the variant.

\begin{figure}[t]
  \centering
  \subfloat[Training Loss]{
    \includegraphics[width=0.225\textwidth]{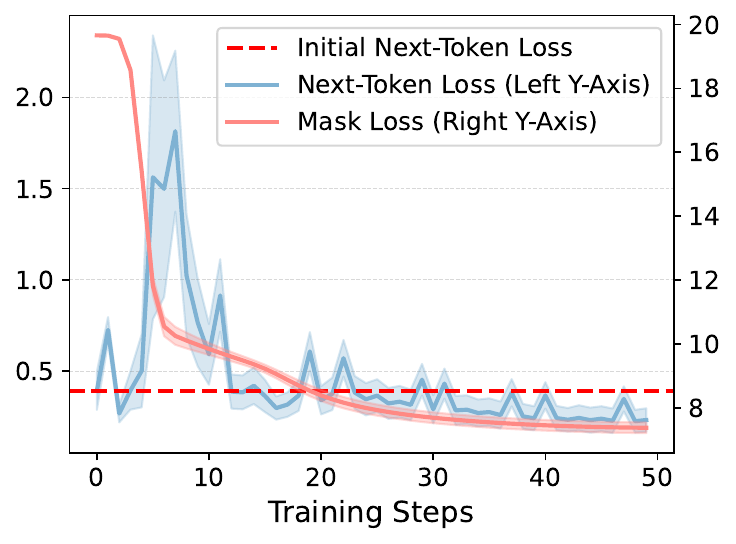}
    \label{fig:mask_training_loss}
  }
  \hfill
  \subfloat[Mask to Accuracy]{
    \includegraphics[width=0.225\textwidth]{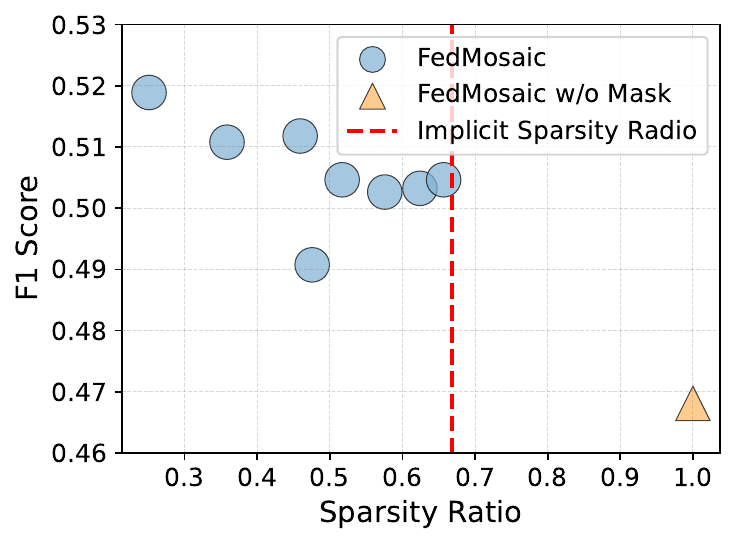}
    \label{fig:sparsity}
  }
  \vspace{-0.5em}
  \caption{Impact of document mask.}
  \vspace{-1.5em}
  \label{fig:mask}
\end{figure}
\subsubsection{Contributions of Document-Specific Mask}
\label{sec:contributions_of_mask}
\figref{fig:mask} demonstrates the role of the document-specific mask in alleviating \textit{intra-silo adapter interference} and its effect on model accuracy. 
We select the number of mask training epochs from $\{3, 5, 10, 20\}$ with varying learning rate.
\figref{fig:mask_training_loss} shows training curves of next-token and mask losses, where both prediction accuracy and mask sparsity improve simultaneously, supporting the hypothesis in \secref{sec:mask}. 
\figref{fig:sparsity} further evaluates accuracy under different sparsity settings, where \sysname consistently outperforms the w/o mask variant. 
Notably, setting $\lambda_{\ell_1}=0$ in \equref{eq:total_loss} drives mask learning solely via next-token loss, yielding implicit sparsity and indicating that document-specific knowledge can be captured by distinct subsets of LoRA parameters.

\subsubsection{Effectiveness of Selective Aggregation}
\label{sec:effectiveness_of_aggregation}
\tabref{tab:agg} presents the effectiveness of selective aggregation on the 2WQA dataset, which is selected for its diversity, under different values of selection $k$.
\sysname outperforms its w/o selection variant as $k$ increases (\eg $20.7\%$ higher in Inf.), demonstrating its ability to filter irrelevant documents and mitigate parameter conflicts.
Moreover, its performance stabilizes with larger $k$, highlighting strong robustness.
\begin{figure*}[t]
  \centering
  
  \begin{minipage}[b]{0.5\textwidth}
    \hspace{2.5mm}
    \includegraphics[width=\textwidth]{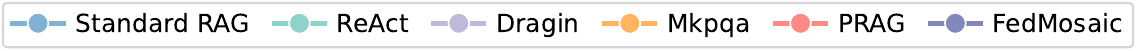}
  \end{minipage}
  \vspace{-3mm} 
  \par
  \subfloat[HQA Compare]{\includegraphics[width=0.16\textwidth]{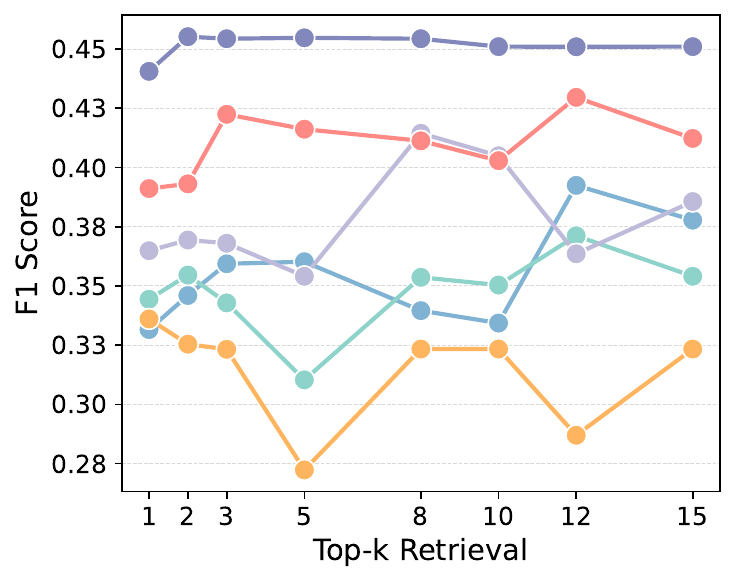}}
  \hfill
  \subfloat[2WQA Bridge]{\includegraphics[width=0.16\textwidth]{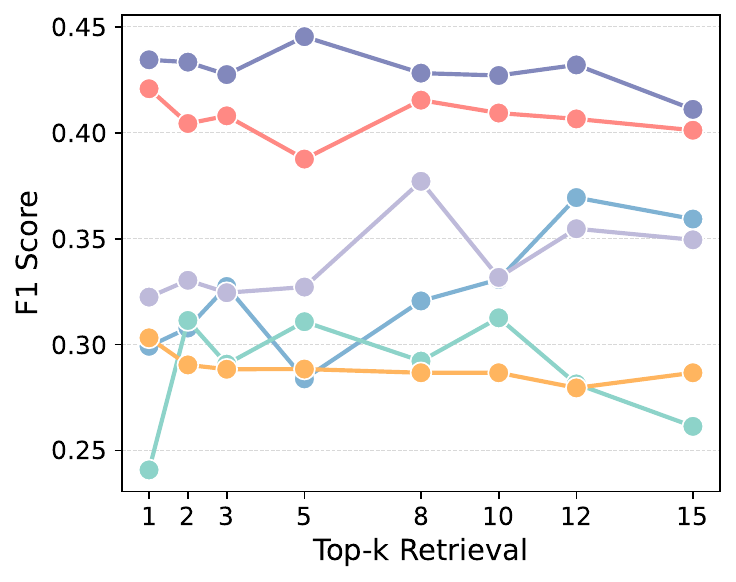}}
  \hfill
  \subfloat[2WQA Inference]{\includegraphics[width=0.16\textwidth]{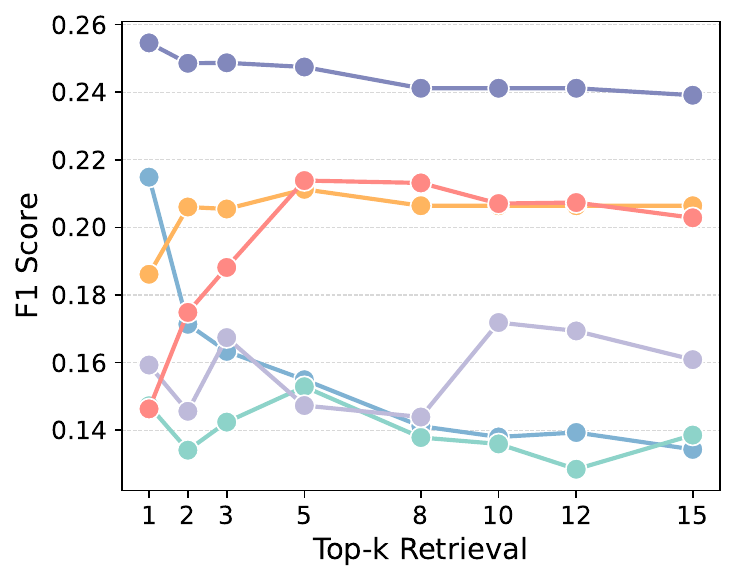}}
  \hfill
  \subfloat[2WQA Compose]{\includegraphics[width=0.16\textwidth]{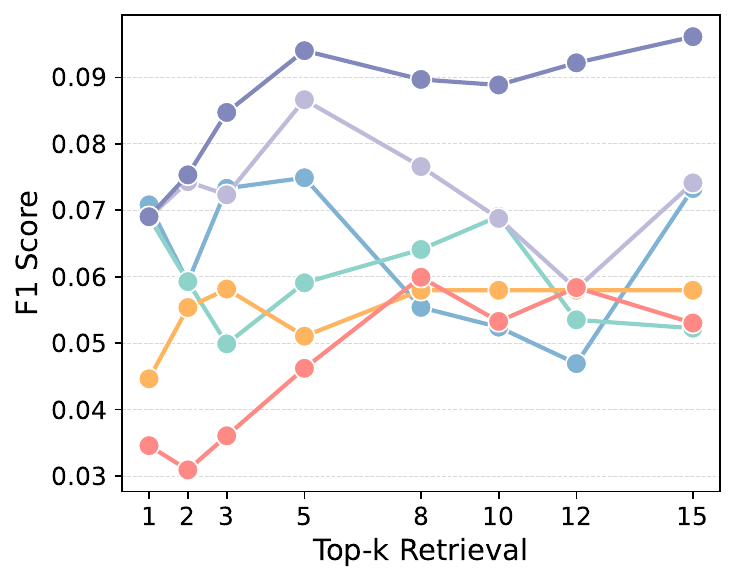}}
  \hfill
  \subfloat[PQA Total]{\includegraphics[width=0.16\textwidth]{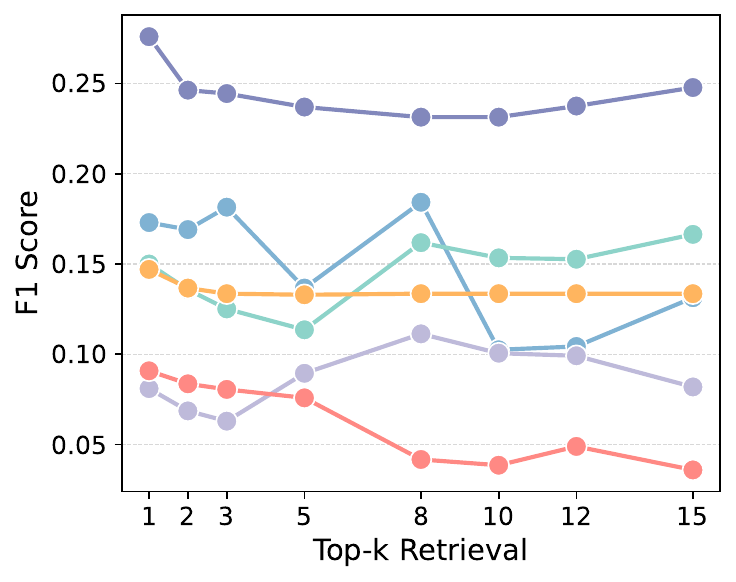}}
  \hfill
  \subfloat[CWQ Total]{\includegraphics[width=0.16\textwidth]{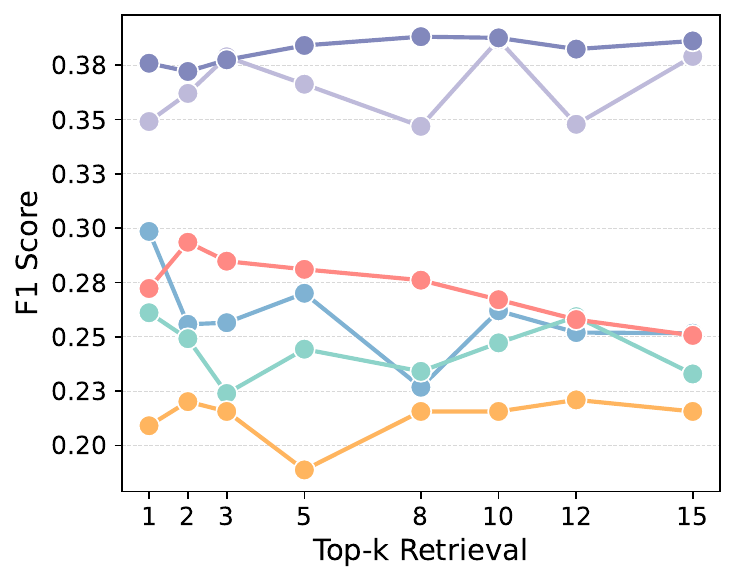}}
  \caption{Performance of federated RAG under varying top-$k$ retrieval settings.}
  \label{fig:top-k}
\end{figure*}

\subsubsection{Impact of Top-$k$ Retrieval}
\label{sec:retrieval_k}
\figref{fig:top-k} compares \sysname with the most competitive FedRAG baseline across all types of datasets under different top-$k$ retrieval settings. Other methods show fluctuating performance, indicating sensitivity to retrieval noise. With selective aggregation, \sysname achieves consistently higher and more stable accuracy, \eg achieving up to a $10.17\%$ average improvement on HQA Compare over the strongest baseline.


\section{Related Work}



\fakeparagraph{Retrieval-Augmented Generation (RAG)}  
RAG \cite{fan2024survey} enhances LLMs with external knowledge, reducing hallucinations and compensating for outdated parametric memory. 
The mainstream is \textit{in-context RAG} \cite{lewis2020retrieval, dong2025understand, guu2020retrieval}, and it can be improved through 
retrieval optimization \cite{robertson2009probabilistic, xu2024list, ouyang2025adarag, lin2024rella}, 
retriever-LLM alignment \cite{shi2025retrieval, cheng2023lift, wang2025unveiling, yu2023augmentation}, and 
multi-round reasoning \cite{yao2023react, su2024dragin, cui2025cirag}. 
However, in-context RAG is unfit for the federated settings by design, since it requires transmitting verbatim documents to the server, which violates the locality constraint.

Parametric RAG \cite{su2025parametric} offers an alternative by compiling knowledge into model parameters rather than context tokens through training.
DyPRAG \cite{tan2025dynamic} trains a parameter translator model to convert documents into parametric knowledge.
By encoding knowledge in reusable parameters rather than prompts, parametric RAG allows documents to remain local while uploading only adapterss, making it suited to federated scenarios.
Our work is built upon parametric RAG, and focuses on the unique efficiency and accuracy challenges when adapting it to federated environments.



\fakeparagraph{Federated Retrieval-Augmented Generation (FedRAG)}
FedRAG extends RAG to settings where knowledge is distributed across silos that cannot share raw documents.  
Recent efforts have explored various strategies for federated search, yet rely on \textit{in-context} RAG.  
For example, FeB4RAG \cite{wang2024feb4rag} provides a benchmark for federated search pipelines,  
MKPQA \cite{shojaee2025federated} enables probabilistic cross-domain search, and 
RAGRoute \cite{guerraoui2025efficient} improves efficiency via query routing. 
Other variants target aggregation or search security.
For example, C-FedRAG \cite{addison2024c} prevent the leakage of raw documents during aggregation and FRAG \cite{zhao2024frag} leverage encrypted similarity search.
Despite these advances, existing frameworks still transmit raw documents across silos, violating locality constraints.

Other efforts adopt privacy-preserving prompt engineering by injecting noise into documents.
For example, DP-Prompt \cite{utpala2023dpprompt} and AUGPE \cite{xie2024differentially} adopt differential privacy strategies.
Sage \cite{zeng2024mitigating} 
adopt anonymization strategies to achieve data desensitization.
However, these approaches suffer severe accuracy degradation in FedRAG settings, limiting their applicability.

Another relevant solution to FedRAG is federated fine-tuning, as explored in FedIT \cite{zhang2024fedit} and FLora \cite{wang2024flora}.  
Compared to FedRAG, federated fine-tuning is less flexible because new documents demand costly retraining and it cannot selectively activate relevant silos.

In contrast, our \sysname introduces the first \textit{federated parametric RAG} framework.  
By encoding documents into local adapters and composing them across silos at inference, \sysname adheres to the locality constraint while providing high flexibility. 
Its optimizations ensure low storage and communication overhead as well as high accuracy, making it a practical and scalable solution for federated knowledge-intensive generation.

\section{Conclusion}

In this work, we explore federated RAG where a central LLM collaborates with distributed knowledge silos without sharing their raw documents. 
We propose \sysname, the first federated RAG framework built on parametric adapters. 
By clustering semantically related documents into multi-document adapters with document-specific masks, \sysname drastically reduces storage and communication overhead while maintaining per-document specificity. 
Furthermore, its selective adapter aggregation mechanism ensures that only relevance-aligned, non-conflicting adapters are combined, mitigating destructive aggregation across silos. 
Extensive experiments on four datasets show that \sysname consistently outperforms state-of-the-art methods across four categories including local RAG, in-context FedRAG, federated fine-tuning, and parametric RAG, while enforcing locality constraints and significantly reducing storage and communication overhead.
We envision \sysname as a foundation for scalable, privacy-preserving knowledge integration, and an important step toward deploying RAG systems in real-world distributed information ecosystems.  



\section*{Acknowledgments}
This work was partially supported by National Science Foundation of China (NSFC) (Grant Nos. 62425202, 62336003), the Beijing Natural Science Foundation (Z230001), and the State Key Laboratory of Complex \& Critical Software Environment (SKLCCSE). 
Zimu Zhou's research is partially supported by the RGC of Hong Kong SAR, China (Project No. CityU 11206425).

\appendix

\section{Proof of NP-Hardness}
\label{sec:np_hard}
We show that the conflict-aware document selection optimization problem is NP-hard by through a reduction from \textsc{CLIQUE} problem \cite{bomze1999maximum}. The proof is given in two steps.

First, We can formally cast the conflict-aware document selection problem as a weighted subgraph selection problem over a complete graph, defined as follows.
\begin{definition}[Weighted Subgraph Selection Problem]
Let $G=(V,E)$ be the complete graph on $n$ vertices. Each vertex $v \in V$ is associated with a positive vertex weight $a_v > 0$, and each unordered pair $\{u,v\} \subseteq V$ is assigned an edge weight $b_{uv} \in \mathbb{R}$. For any subset $S \subseteq V$ with cardinality $|S| = k$, the total weight of $S$ is defined as
\begin{equation}
    W(S) \;=\; \sum_{v \in S} a_v \;- \sum_{\{u,v\} \subseteq S} b_{uv}.
\end{equation}
The Weighted Subgraph Selection Problem asks: given a weighted complete graph with vertex weights $\{a_v\}_{v \in V}$, edge weights $\{b_{uv}\}_{\{u,v\} \subseteq V}$, and an integer $k$, find a subset $S \subseteq V$ of size $k$ that maximizes $W(S)$.
\end{definition}

Then, we reduce from the classical \textsc{CLIQUE} problem, which is known to be NP-complete \cite{bomze1999maximum}. Given a graph 
$G'=(V',E')$ and an integer $q$, the \textsc{CLIQUE} problem asks whether there exists a subset 
$C \subseteq V'$ with $|C|=q$ such that every pair of vertices in $C$ is connected by an edge in $E'$.
From an instance $(G',q)$ we construct an instance of the Weighted Subgraph Selection Problem as follows:
\begin{itemize}
    \item The vertex set of the new instance is $V := V'$.
    \item For each $v \in V$, set the vertex weight $a_v := 1$.
    \item For each unordered pair $\{u,v\} \subseteq V$, define the edge weight
    \begin{equation}
        b_{uv} \;=\;
        \begin{cases}
            0, & \{u,v\} \in E', \\
            B, & \{u,v\} \notin E',
        \end{cases}
    \end{equation}
    where $B$ is a sufficiently large constant (e.g., $B > q$).
    \item The target subset size is set to $k := q$.
\end{itemize}

Consider any subset $S \subseteq V$ with $|S|=q$. Its weight is $W(S) = q - B \cdot t(S)$, 
where $t(S)$ denotes the number of non-edges induced by $S$, i.e.,
\begin{equation}
    t(S) \;=\; \big|\{\{u,v\}\subseteq S : \{u,v\} \notin E'\}\big|.
\end{equation}
By construction, if $S$ is a clique of size $q$, then $t(S)=0$ and $W(S)=q$. Otherwise, $t(S)\ge 1$ and thus $W(S) \;\le\; q-B$.
Since $B>q$, the global maximum of $W(S)$ equals $q$ if and only if $G'$ contains a clique of size $q$. 
Hence, deciding whether $G'$ admits a clique of size $q$ can be answered by solving the weighted subgraph selection problem on the constructed instance. 
This establishes a polynomial-time reduction 
\begin{equation}
    \textsc{CLIQUE} \;\leq_p\; \textsc{Weighted Subgraph Selection}.
\end{equation}
Since \textsc{CLIQUE} is NP-complete, it follows immediately that the weighted subgraph selection problem, and consequently the conflict-aware document selection problem, is NP-hard.

\section{Supplementary Experiment}

\subsection{Experimental Settings}
\subsubsection{Experiment Configuration}
\label{app:config}
We conduct our experiments on a machine equipped with an Intel Xeon Gold 6230R CPU and four NVIDIA A100 GPUs, each providing 40 GB of memory for computation. 
The fundamental experimental configurations of \sysname adopt mask training for 10 epochs with learning rate $3 \times 10^{-1}$, $\lambda_{l1}=10^{-4}$, $\lambda_{ol}=0.8$, $C \in \{5,8\}$, and LoRA training for 3 or 5 epochs.  The setups for document augmentation and adapter training follow \cite{su2025parametric}.

\subsubsection{Hyperparameters}
We set the default retrieval to the top-5 documents per silo for each query. 
Other hyperparameters of the baselines are detailed below.
\begin{itemize}
    \item \textbf{CoTRAG} \cite{wei2022chain}: 
    We set the few-shot number to 11. 
    \item \textbf{ReAct} \cite{yao2023react}: 
    We set the maximum attempts per step to 10. 
    \item \textbf{MKPQA} \cite{shojaee2025federated}:  
    We set the threshold to $\tau=400$. 
    \item \textbf{RAGRoute} \cite{guerraoui2025efficient}: 
    We train the router on 300 generated samples for 150 epochs. 
    \item \textbf{FedIT} \cite{zhang2024fedit}, \textbf{FLora} \cite{wang2024flora}: 
    We perform 10 local epochs per silo and 3 rounds of server aggregation. 
    \item \textbf{Dragin} \cite{su2024dragin}, \textbf{PRAG} \cite{su2025parametric}: 
    We use the default hyperparameters from the original papers. 
\end{itemize}

For in-context integration methods, we design distinct prompts for the silo side and the server side, since silos only upload candidate answers to the server.

\subsection{Additional Experiments}
\subsubsection{Impact of Balanced Document Clustering}
\label{app:clustering}
\figref{fig:clustering} evaluates balanced document clustering on four datasets at $c=5,15$, reporting the average hit ratio (\ie the proportion of top-$3$ documents in the same cluster) and the cluster size deviation. 
Compared with random clustering, the balanced strategy more effectively controls cluster sizes and ensures stability across different datasets, \eg reducing the deviation by $85.2\%$ on HQA at $c=15$. 
It also consistently groups semantically related documents into the same adapter, thereby increasing the likelihood that retrieved candidates share an adapter and significantly lowering communication cost, \eg the average top-$3$ hit ratio improves from $0.352$ to $0.607$ in CWQ at $c=5$.

\begin{figure}[b]
  \centering
  \subfloat[Communication Reduction $c=5$]{
    \includegraphics[width=0.22\textwidth]{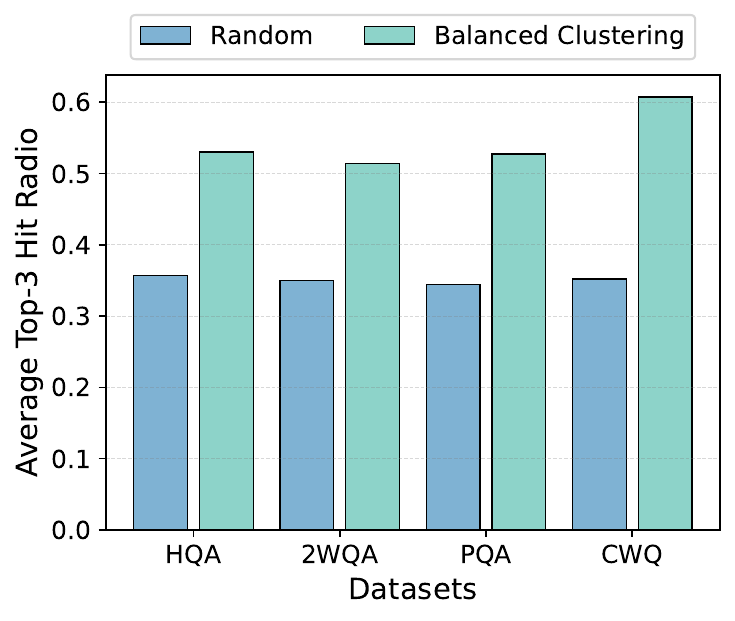}
    \label{fig:cluster_c=5}
  }
  \hfill
  \subfloat[Cluster Standard Deviation $c=5$]{
    \includegraphics[width=0.22\textwidth]{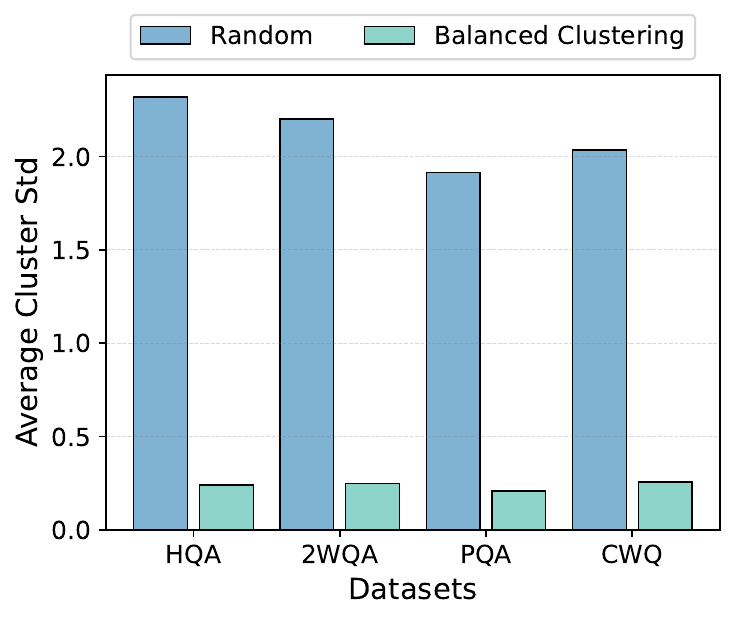}
    \label{fig:cluster_std_c=5}
  }

  \subfloat[Communication Reduction $c=15$]{
    \includegraphics[width=0.22\textwidth]{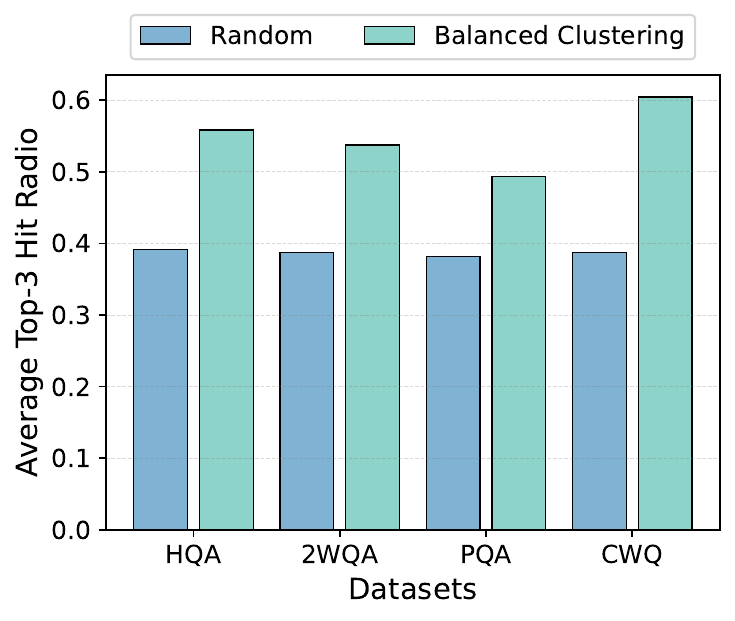}
    \label{fig:cluster_c=15}
  }
  \hfill
  \subfloat[Cluster Standard Deviation $c=15$]{
    \includegraphics[width=0.22\textwidth]{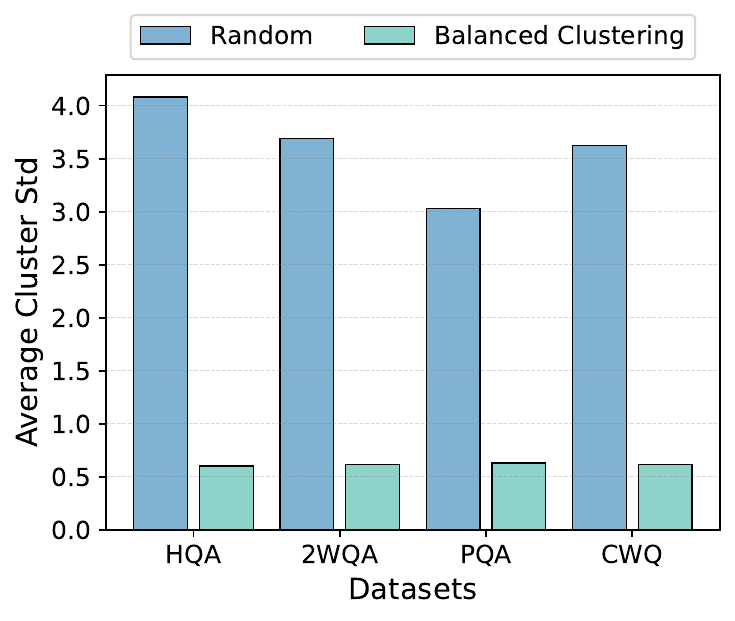}
    \label{fig:cluster_std_c=15}
  }
  \caption{Balanced and random document clustering.}
  \label{fig:clustering}
\end{figure}
\begin{figure}[b]
  \centering
  \subfloat[$\lambda_{\ell_1}=10^{-5}$]{
    \includegraphics[width=0.225\textwidth]{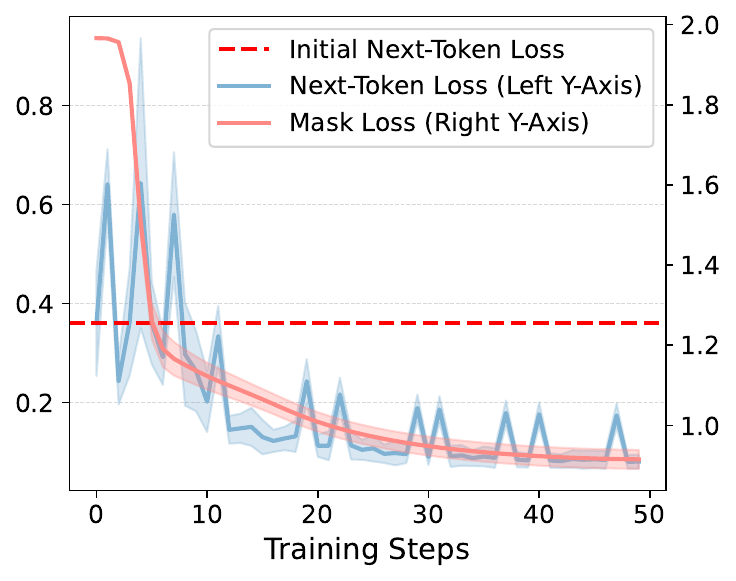}
    \label{fig:loss_lambda=1e-5}
  }
  \hfill
  \subfloat[$\lambda_{\ell_1}=10^{-6}$]{
    \includegraphics[width=0.225\textwidth]{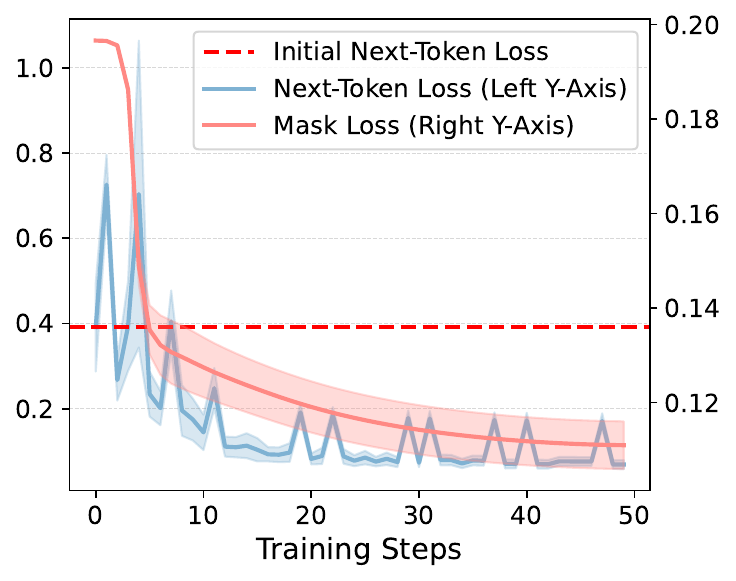}
    \label{fig:loss_lambda=1e-6}
  }
  \caption{Mask training loss under varying $\lambda_{\ell_1}$.}
  \label{fig:lanbda_l1}
\end{figure}
\subsubsection{Impact of balance coefficient $\lambda_{\ell_1}$}
\figref{fig:lanbda_l1} reports the mask training loss curves obtained under different settings of $\lambda_{\ell_1}$. 
Across varying $\lambda_{\ell_1}$, both the next-token loss and the mask loss exhibit substantial reductions throughout the training process.
Meanwhile, larger $\lambda_{\ell_1}$ values yield sparser masks, effectively enabling a trade-off between training accuracy and mask sparsity.

\balance

\clearpage

\bibliographystyle{ACM-Reference-Format}
\balance
\bibliography{ref}

@inproceedings{su2024dragin,
    title = {DRAGIN: Dynamic retrieval augmented generation based on the real-time information needs of large language models},
    author = {Su, Weihang  and Tang, Yichen  and Ai, Qingyao  and Wu, Zhijing  and Liu, Yiqun},
    booktitle = {ACL},
    pages = {12991--13013},
    year = {2024},
}

@inproceedings{mallen2023popqa,
  title={When not to trust language models: Investigating effectiveness of parametric and non-parametric memories},
  author={Mallen, Alex and Asai, Akari and Zhong, Victor and Das, Rajarshi and Khashabi, Daniel and Hajishirzi, Hannaneh},
  booktitle = {ACL},
  pages={9802--9822},
  year={2023}
}

@inproceedings{yu2023augmentation,
  title={Augmentation-adapted retriever improves generalization of language models as generic plug-in},
  author={Yu, Zichun and Xiong, Chenyan and Yu, Shi and Liu, Zhiyuan},
  booktitle = {ACL},
  pages={2421--2436},
  year={2023}
}

@inproceedings{zeng2024good,
  title={The good and the bad: Exploring privacy issues in retrieval-augmented generation (RAG)},
  author={Zeng, Shenglai and Zhang, Jiankun and He, Pengfei and Liu, Yiding and Xing, Yue and Xu, Han and Ren, Jie and Chang, Yi and Wang, Shuaiqiang and Yin, Dawei and others},
  booktitle={Findings of ACL},
  pages={4505--4524},
  year={2024}
}

@inproceedings{chen2025omniRAG,
    title = {Towards omni-RAG: Comprehensive retrieval-augmented generation for large language models in medical applications},
    author = {Chen, Zhe  and Liao, Yusheng  and Jiang, Shuyang  and Wang, Pingjie  and Guo, YiQiu  and Wang, Yanfeng  and Wang, Yu},
    booktitle = {ACL},
    pages = {15285--15309},
    year = {2025},
}

@inproceedings{lee2025shifting,
  title={Shifting from Ranking to Set Selection for Retrieval Augmented Generation},
  author={Lee, Dahyun and Jo, Yongrae and Park, Haeju and Lee, Moontae},
    booktitle = {ACL},
  pages={17606--17619},
  year={2025}
}

@inproceedings{du2025neural,
    title = {Neural Parameter Search for Slimmer Fine-Tuned Models and Better Transfer},
    author = {Du, Guodong  and Fang, Zitao  and Li, Jing  and Li, Junlin  and Jiang, Runhua  and Yu, Shuyang  and Guo, Yifei  and Chen, Yangneng  and Goh, Sim Kuan  and Tang, Ho-Kin  and He, Daojing  and Liu, Honghai  and Zhang, Min},
    booktitle = {ACL},
    year = {2025},
    pages = {32668--32687},
}

@inproceedings{shojaee2025federated,
  title={Federated retrieval augmented generation for multi-product question answering},
  author={Shojaee, Parshin and Harsha, Sai Sree and Luo, Dan and Maharaj, Akash and Yu, Tong and Li, Yunyao},
  booktitle={COLING},
  volume={Industry Track},
  pages={387--397},
  year={2025}
}

@inproceedings{ho2020wikiqa,
  title={Constructing a multi-hop qa dataset for comprehensive evaluation of reasoning steps},
  author={Ho, Xanh and Nguyen, Anh-Khoa Duong and Sugawara, Saku and Aizawa, Akiko},
  booktitle={COLING},
  pages={6609--6625},
  year={2020}
}

@inproceedings{yang2018hotpotqav,
  title={HotpotQA: A dataset for diverse, explainable multi-hop question answering},
  author={Yang, Zhilin and Qi, Peng and Zhang, Saizheng and Bengio, Yoshua and Cohen, William and Salakhutdinov, Ruslan and Manning, Christopher D},
  booktitle={EMNLP},
  pages={2369--2380},
  year={2018}
}

@inproceedings{utpala2023dpprompt,
  title={Locally differentially private document generation using zero shot prompting},
  author={Utpala, Saiteja and Hooker, Sara and Chen, Pin-Yu},
  booktitle={Findings of EMNLP},
  year={2023}
}

@inproceedings{zhang2024fedit,
  title={Towards building the federatedgpt: Federated instruction tuning},
  author={Zhang, Jianyi and Vahidian, Saeed and Kuo, Martin and Li, Chunyuan and Zhang, Ruiyi and Yu, Tong and Wang, Guoyin and Chen, Yiran},
  booktitle={ICASSP},
  pages={6915--6919},
  year={2024}
}

@inproceedings{yao2023react,
  title={React: Synergizing reasoning and acting in language models},
  author={Yao, Shunyu and Zhao, Jeffrey and Yu, Dian and Du, Nan and Shafran, Izhak and et al},
  booktitle={ICLR},
  year={2023}
}

@inproceedings{hu2022lora,
  title={LoRA: Low-rank adaptation of large language models},
  author={Hu, Edward J and Wallis, Phillip and Allen-Zhu, Zeyuan and Li, Yuanzhi and Wang, Shean and Wang, Lu and Chen, Weizhu and others},
  booktitle={ICLR},
  year={2022}
}

@inproceedings{carlini2022quantifying,
  title={Quantifying memorization across neural language models},
  author={Carlini, Nicholas and Ippolito, Daphne and Jagielski, Matthew and Lee, Katherine and Tramer, Florian and Zhang, Chiyuan},
  booktitle={ICLR},
  year={2022}
}

@inproceedings{guu2020retrieval,
  title={Retrieval augmented language model pre-training},
  author={Guu, Kelvin and Lee, Kenton and Tung, Zora and Pasupat, Panupong and Chang, Mingwei},
  booktitle={ICML},
  pages={3929--3938},
  year={2020}
}

@inproceedings{yu2024dare,
  title={Language models are super mario: Absorbing abilities from homologous models as a free lunch},
  author={Yu, Le and Yu, Bowen and Yu, Haiyang and Huang, Fei and Li, Yongbin},
  booktitle={ICML},
  year={2024}
}

@inproceedings{xu2024winning,
  title={Random masking finds winning tickets for parameter efficient fine-tuning},
  author={Xu, Jing and Zhang, Jingzhao},
  booktitle={ICML},
  pages={55501--55524},
  year={2024},
}

@inproceedings{xie2024differentially,
  title={Differentially private synthetic data via foundation model apis 2: Text},
  author={Xie, Chulin and Lin, Zinan and Backurs, Arturs and Gopi, Sivakanth and Yu, Da and Inan, Huseyin A and Nori, Harsha and Jiang, Haotian and Zhang, Huishuai and Lee, Yin Tat and others},
  booktitle={ICML},
  year={2024}
}

@inproceedings{talmor2018complexqa,
  title={The web as a knowledge-base for answering complex questions},
  author={Talmor, Alon and Berant, Jonathan},
  booktitle={NAACL},
  pages={641--651},
  year={2018}
}

@article{lewis2020retrieval,
  title={Retrieval-augmented generation for knowledge-intensive nlp tasks},
  author={Lewis, Patrick and Perez, Ethan and Piktus, Aleksandra and Petroni, Fabio and Karpukhin, Vladimir and Goyal, Naman and K{\"u}ttler, Heinrich and Lewis, Mike and Yih, Wen-tau and Rockt{\"a}schel, Tim and others},
  journal={NeurIPS},
  volume={33},
  pages={9459--9474},
  year={2020}
}

@article{wei2022chain,
  title={Chain-of-thought prompting elicits reasoning in large language models},
  author={Wei, Jason and Wang, Xuezhi and Schuurmans, Dale and Bosma, Maarten and Xia, Fei and Chi, Ed and Le, Quoc V and Zhou, Denny and others},
  journal={NeurIPS},
  volume={35},
  pages={24824--24837},
  year={2022}
}

@article{wang2024flora,
  title={Flora: Federated fine-tuning large language models with heterogeneous low-rank adaptations},
  author={Wang, Ziyao and Shen, Zheyu and He, Yexiao and Sun, Guoheng and Wang, Hongyi and Lyu, Lingjuan and Li, Ang},
  journal={NeurIPS},
  volume={37},
  pages={22513--22533},
  year={2024}
}

@article{cheng2023lift,
  title={Lift yourself up: Retrieval-augmented text generation with self-memory},
  author={Cheng, Xin and Luo, Di and Chen, Xiuying and Liu, Lemao and Zhao, Dongyan and Yan, Rui},
  journal={NeurIPS},
  volume={36},
  pages={43780--43799},
  year={2023}
}

@article{lu2024twin,
  title={Twin-merging: Dynamic integration of modular expertise in model merging},
  author={Lu, Zhenyi and Fan, Chenghao and Wei, Wei and Qu, Xiaoye and Chen, Dangyang and Cheng, Yu},
  journal={NeurIPS},
  volume={37},
  pages={78905--78935},
  year={2024}
}

@article{huang2024emr,
  title={Emr-merging: Tuning-free high-performance model merging},
  author={Huang, Chenyu and Ye, Peng and Chen, Tao and He, Tong and Yue, Xiangyu and Ouyang, Wanli},
  journal={NeurIPS},
  volume={37},
  pages={122741--122769},
  year={2024}
}

@article{ortiz2023task,
  title={Task arithmetic in the tangent space: Improved editing of pre-trained models},
  author={Ortiz-Jimenez, Guillermo and Favero, Alessandro and Frossard, Pascal},
  journal={NeurIPS},
  volume={36},
  pages={66727--66754},
  year={2023}
}

@inproceedings{su2025parametric,
  title={Parametric retrieval augmented generation},
  author={Su, Weihang and Tang, Yichen and Ai, Qingyao and Yan, Junxi and Wang, Changyue and Wang, Hongning and Ye, Ziyi and Zhou, Yujia and Liu, Yiqun},
  booktitle={SIGIR},
  pages={1240--1250},
  year={2025}
}

@inproceedings{wang2024feb4rag,
  title={Feb4rag: Evaluating federated search in the context of retrieval augmented generation},
  author={Wang, Shuai and Khramtsova, Ekaterina and Zhuang, Shengyao and Zuccon, Guido},
  booktitle={SIGIR},
  pages={763--773},
  year={2024}
}

@inproceedings{cui2025cirag,
  title={CIRAG: Retrieval-Augmented Language Model with Collective Intelligence},
  author={Cui, Chenxu and Fan, Haihui and Zhang, Jinchao and Shen, Lin and Li, Bo and Wang, Weiping},
  booktitle={SIGIR},
  pages={1316--1326},
  year={2025}
}

@inproceedings{shi2025retrieval,
  title={Retrieval Augmented Generation with Collaborative Filtering for Personalized Text Generation},
  author={Shi, Teng and Xu, Jun and Zhang, Xiao and Zang, Xiaoxue and Zheng, Kai and Song, Yang and Li, Han},
  booktitle={SIGIR},
  year={2025}
}

@inproceedings{wang2025unveiling,
  title={Unveiling Knowledge Utilization Mechanisms in LLM-based Retrieval-Augmented Generation},
  author={Wang, Yuhao and Ren, Ruiyang and Wang, Yucheng and Zhao, Wayne Xin and Liu, Jing and Wu, Hua and Wang, Haifeng},
  booktitle={SIGIR},
  year={2025}
}

@inproceedings{fan2024survey,
  title={A survey on rag meeting llms: Towards retrieval-augmented large language models},
  author={Fan, Wenqi and Ding, Yujuan and Ning, Liangbo and Wang, Shijie and Li, Hengyun and Yin, Dawei and Chua, Tat-Seng and Li, Qing},
  booktitle={SIGKDD},
  pages={6491--6501},
  year={2024}
}

@inproceedings{dong2025understand,
  title={Understand what LLM needs: Dual preference alignment for retrieval-augmented generation},
  author={Dong, Guanting and Zhu, Yutao and Zhang, Chenghao and Wang, Zechen and Wen, Ji-Rong and Dou, Zhicheng},
  booktitle={WWW},
  pages={4206--4225},
  year={2025}
}

@inproceedings{zhao2025medrag,
  title={Medrag: Enhancing retrieval-augmented generation with knowledge graph-elicited reasoning for healthcare copilot},
  author={Zhao, Xuejiao and Liu, Siyan and Yang, Su-Yin and Miao, Chunyan},
  booktitle={WWW},
  pages={4442--4457},
  year={2025}
}

@inproceedings{xu2024list,
  title={List-aware reranking-truncation joint model for search and retrieval-augmented generation},
  author={Xu, Shicheng and Pang, Liang and Xu, Jun and Shen, Huawei and Cheng, Xueqi},
  booktitle={WWW},
  pages={1330--1340},
  year={2024}
}

@inproceedings{lin2024rella,
  title={Rella: Retrieval-enhanced large language models for lifelong sequential behavior comprehension in recommendation},
  author={Lin, Jianghao and Shan, Rong and Zhu, Chenxu and Du, Kounianhua and Chen, Bo and Quan, Shigang and Tang, Ruiming and Yu, Yong and Zhang, Weinan},
  booktitle={WWW},
  pages={3497--3508},
  year={2024}
}

@inproceedings{guerraoui2025efficient,
  title={Efficient federated search for retrieval-augmented generation},
  author={Guerraoui, Rachid and Kermarrec, Anne-Marie and Petrescu, Diana and Pires, Rafael and Randl, Mathis and de Vos, Martijn},
  booktitle={Proceedings of the 5th Workshop on Machine Learning and Systems},
  pages={74--81},
  year={2025}
}

@inproceedings{ouyang2025adarag,
  title={AdaRAG: Adaptive Optimization for Retrieval Augmented Generation with Multilevel Retrievers at the Edge},
  author={Ouyang, Tao and Hong, Guihang and Zhao, Kongyange and Zhou, Zhi and Wu, Weigang and Lv, Zhaobiao and Chen, Xu},
  booktitle={INFOCOM},
  pages={1--10},
  year={2025},
  organization={IEEE}
}

@article{shen2025rapid,
  title={Rapid deployment of large language model DeepSeek in Chinese hospitals demands a regulatory response},
  author={Shen, Tianyi and Li, Yuxi and Cao, Yanlin and Du, Xin and Wang, Xinru and Zhang, Yajuan and Zhang, Yi},
  journal={Nature Medicine},
  pages={1--6},
  year={2025}
}

@article{boycott2025rdi,
  title={The RDI--Lancet Commission on Rare Diseases: improving visibility to address health-care disparities for 400 million people},
  author={Boycott, Kym M and Giugliani, Roberto},
  journal={The Lancet},
  volume={405},
  number={10479},
  pages={605--607},
  year={2025}
}

@article{bradley2000constrained,
  title={Constrained k-means clustering},
  author={Bradley, Paul S and Bennett, Kristin P and Demiriz, Ayhan},
  journal={Microsoft Research, Redmond},
  volume={20},
  year={2000}
}

@article{addison2024c,
  title={C-fedrag: A confidential federated retrieval-augmented generation system},
  author={Addison, Parker and Nguyen, Minh-Tuan H and Medan, Tomislav and Shah, Jinali and Manzari, Mohammad T and McElrone, Brendan and Lalwani, Laksh and More, Aboli and Sharma, Smita and Roth, Holger R and others},
  journal={arXiv preprint arXiv:2412.13163},
  year={2024}
}

@article{zhao2024frag,
  title={Frag: Toward federated vector database management for collaborative and secure retrieval-augmented generation},
  author={Zhao, Dongfang},
  journal={arXiv preprint arXiv:2410.13272},
  year={2024}
}

@article{tan2025dynamic,
  title={Dynamic parametric retrieval augmented generation for test-time knowledge enhancement},
  author={Tan, Yuqiao and He, Shizhu and Liao, Huanxuan and Zhao, Jun and Liu, Kang},
  journal={arXiv preprint arXiv:2503.23895},
  year={2025}
}

@article{zeng2024mitigating,
  title={Mitigating the privacy issues in retrieval-augmented generation (rag) via pure synthetic data},
  author={Zeng, Shenglai and Zhang, Jiankun and He, Pengfei and Ren, Jie and Zheng, Tianqi and Lu, Hanqing and Xu, Han and Liu, Hui and Xing, Yue and Tang, Jiliang},
  journal={arXiv preprint arXiv:2406.14773},
  year={2024}
}

@misc{EU2016GDPR,
  title        = {Regulation (EU) 2016/679 of the European Parliament and of the Council of 27 April 2016 on the protection of natural persons with regard to the processing of personal data and on the free movement of such data, and repealing Directive 95/46/EC (General Data Protection Regulation)},
  author       = {European Union},
  year         = {2016},
  howpublished = {https://eur-lex.europa.eu/eli/reg/2016/679/oj/eng},
  note         = {Accessed: 2025-9-19}
}

@misc{US1996HIPAA,
  title        = {Health Insurance Portability and Accountability Act of 1996},
  author       = {U.S. Congress},
  year         = {1996},
  howpublished = {https://www.govinfo.gov/content/pkg/PLAW-104publ191/pdf/PLAW-104publ191.pdf},
  note         = {Accessed: 2025-9-19}
}

@article{robertson2009probabilistic,
  title={The probabilistic relevance framework: BM25 and beyond},
  author={Robertson, Stephen and Zaragoza, Hugo and others},
  journal={Foundations and Trends{\textregistered} in Information Retrieval},
  volume={3},
  pages={333--389},
  year={2009},
}

@incollection{bomze1999maximum,
  title={The maximum clique problem},
  author={Bomze, Immanuel M and Budinich, Marco and Pardalos, Panos M and Pelillo, Marcello},
  booktitle={Handbook of Combinatorial Optimization: Supplement Volume A},
  pages={1--74},
  year={1999},
  publisher={Springer}
}
    


\end{document}